\documentclass[conference]{IEEEtran}

\usepackage[numbers]{natbib}
\usepackage{multicol}
\usepackage{amssymb}
\usepackage{subcaption}
\usepackage[colorlinks=true, linkcolor=blue, urlcolor=red, citecolor=green]{hyperref}
\usepackage{graphics,graphicx,caption}
\usepackage{amsmath}
\usepackage{pifont}
\usepackage{multirow}
\usepackage{booktabs} 

\begin{document}

\title{
Dexonomy: Synthesizing All Dexterous \\ Grasp Types in a Grasp Taxonomy
}




%
\author{\authorblockN{Jiayi Chen$^{1,2*}$,
Yubin Ke$^{1,2*}$,
Lin Peng$^{2}$, 
He Wang$^{1,2,3\dagger}$}
\authorblockA{$^{1}$Peking University, $^2$Galbot, $^3$Beijing Academy of Artificial Intelligence}
\authorblockA{$^*$Equal contribution, $^\dagger$Corresponding author}
}

\makeatletter
\let\@oldmaketitle\@maketitle
\renewcommand{\@maketitle}{\@oldmaketitle
\centering
  \includegraphics[width=\linewidth]{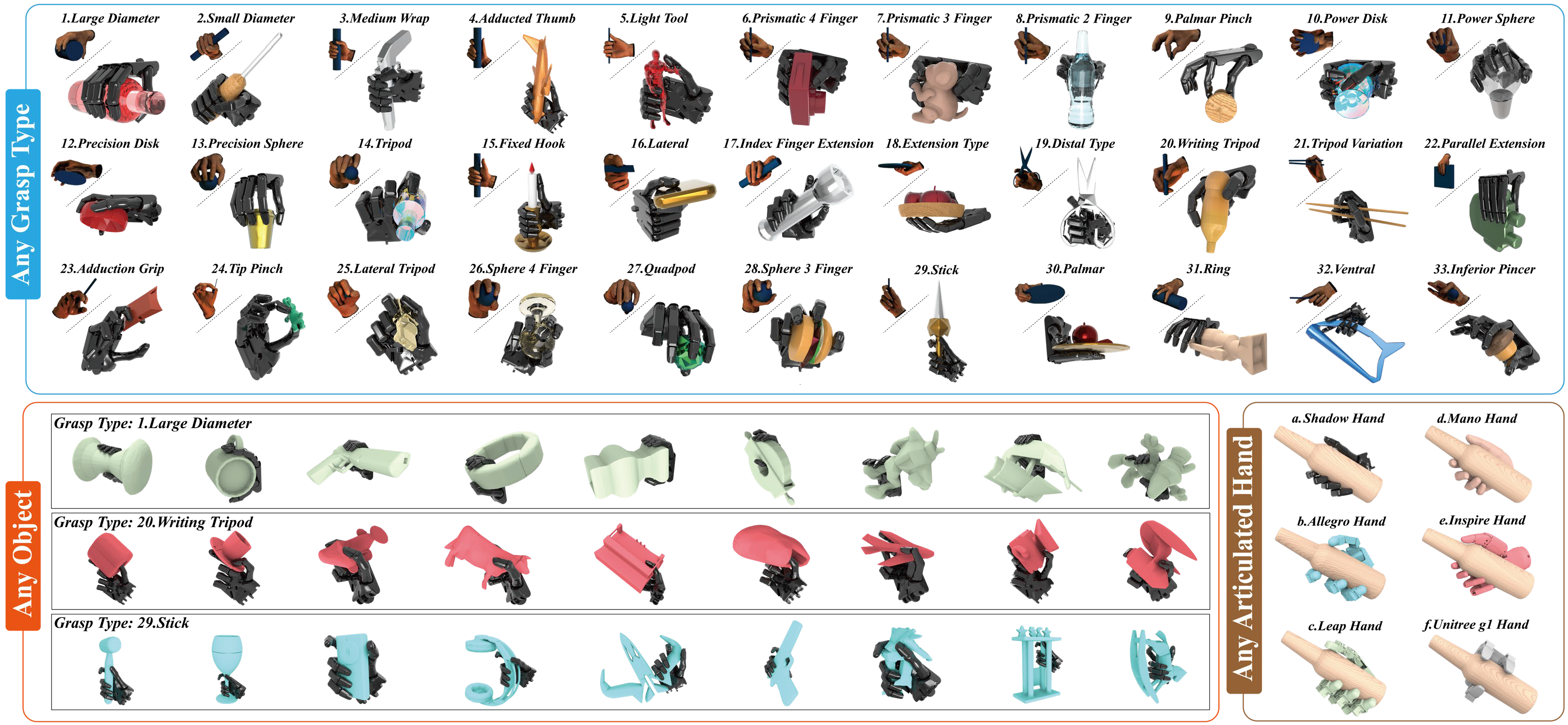}
  \captionof{figure}{For \textbf{any grasp type} in the GRASP taxonomy~\cite{feix2015grasp}, \textbf{any object}, and \textbf{any articulated hand}, our pipeline efficiently synthesizes contact-rich, penetration-free, and physically plausible dexterous grasps, starting from only one human-annotated grasp template to specify an initial hand pose and contact information per hand and grasp type. 
  }
  \label{fig:teaser}
  }
\makeatother

\maketitle
\addtocounter{figure}{-1}

\begin{abstract}
Generalizable dexterous grasping with suitable grasp types is a fundamental skill for intelligent robots. Developing such skills requires a large-scale and high-quality dataset that covers numerous grasp types (i.e., at least those categorized by the GRASP taxonomy), but collecting such data is extremely challenging. Existing automatic grasp synthesis methods are often limited to specific grasp types or object categories, hindering scalability.
This work proposes an efficient pipeline capable of synthesizing contact-rich, penetration-free, and physically plausible grasps for any grasp type, object, and articulated hand. Starting from a single human-annotated template for each hand and grasp type, our pipeline tackles the complicated synthesis problem with two stages: optimize the object to fit the hand template first, and then locally refine the hand to fit the object in simulation. To validate the synthesized grasps, we introduce a contact-aware control strategy that allows the hand to apply the appropriate force at each contact point to the object. Those validated grasps can also be used as new grasp templates to facilitate future synthesis.
Experiments show that our method significantly outperforms previous type-unaware grasp synthesis baselines in simulation. Using our algorithm, we construct a dataset containing 10.7k objects and 9.5M grasps, covering 31 grasp types in the GRASP taxonomy. Finally, we train a type-conditional generative model that successfully performs the desired grasp type from single-view object point clouds, achieving an $82.3\%$ success rate in real-world experiments. Project page: \href{https://pku-epic.github.io/Dexonomy}{https://pku-epic.github.io/Dexonomy}.

\end{abstract}

\IEEEpeerreviewmaketitle

\section{Introduction}

Dexterous grasping is a fundamental skill for intelligent robots, enabling flexible interaction with the environment. However, this potential remains largely under-explored. Although dexterous grasping has received increasing attention, most prior work focuses on whether a dexterous hand can successfully grasp an object, rather than considering \textit{how} to grasp it. As a result, the dexterous hand loses its dexterity and becomes functionally similar to a large parallel gripper. True dexterous grasping is not merely about ``grasping with dexterous hands", but about ``grasping dexterously with appropriate grasp types based on the task requirement". For example, when a robot needs to securely grasp an apple or hold a knife to cut, it should use a power grasp to envelop the object. Conversely, when grasping a lightweight or flat object from the table, a precision grasp using the fingertips would be more suitable.

To develop such intelligent skills, there are two key challenges: (1) selecting the appropriate grasp type based on the task and (2) generating high-quality grasps for specified types and objects. The first challenge is a high-level decision-making problem and can take advantage of recent advances in large vision-language models, e.g., GPT-4o~\cite{hurst2024gpt}, as a temporary solution. However, the second challenge is less studied and represents a significant bottleneck, which is the main focus of this paper. To address the problem of type-aware grasp synthesis with data-driven methods, the first step is to build a large-scale grasp dataset that at least includes most of the grasp types in the GRASP taxonomy~\cite{feix2015grasp}. However, collecting and annotating grasp data, particularly for multi-fingered hands in contact-rich scenarios, remains a big challenge.

Several approaches have been explored for automatically synthesizing a large-scale dexterous grasp dataset, but most of them suffer from various limitations. Analytical grasp synthesis methods~\cite{wang2023dexgraspnet, turpin2023fast, li2023frogger, chen2024springgrasp, chen2024bodex} are often applicable to any object, but most of them are type-unaware and the synthesized grasps only belong to limited types. This is because specifying flexible grasp types solely through analytical metrics is challenging. Moreover, these methods often produce unnatural hand poses, as they prioritize \textit{force closure}, which does not always align with human habits. Another line of research~\cite{yang2022oakink, wei2024learning, wu2024cross} focuses on transferring functional dexterous grasps by mapping object contact regions. While these methods generate more human-like grasps and support a wider range of grasp types, they are limited to objects that are geometrically similar or axis-aligned with the initial demonstration, making them less scalable.

In this work, we propose a novel pipeline based on sampling and optimization to address these challenges. As shown in Figure~\ref{fig:teaser}, our algorithm can efficiently synthesize high-quality dexterous grasps for any grasp type, object, and articulated hand, requiring only one human-annotated grasp template per hand and grasp type. Our synthesized grasps achieve rich hand-object contact (e.g., $>10$ hand links within 2 mm of the object for power grasps), guarantee penetration-free poses via collision mesh verification, and satisfy force closure under six-axis testing in MuJoCo~\cite{todorov2012mujoco} — all with shared hyperparameters across grasp types, objects, and hands.

Our key insight is that grasping can be framed as a geometric matching problem, where the hand and object should align through contact points. We begin by introducing a human-annotated grasp template that specifies the initial hand pose and contact information (i.e., points and normals). Unlike previous methods that directly adjust the hand pose to fit the object, we first use a lightweight stage that samples and optimizes the object pose to match the hand contacts defined in the grasp template. This stage supports hundreds of thousands of initial samples processed in parallel on a single GPU and leaves only a small number of promising results for the next stage. The combination of dense sampling and optimization helps avoid local optima without sacrificing efficiency. 

After aligning the object pose, the hand only needs a slight refinement to get a good grasp. This dual-stage design not only eases the hand refinement stage, but also ensures that the final grasp remains similar to the initial grasp template and thus remains natural. In contrast to most prior work~\cite{jiang2021hand, wang2023dexgraspnet, chen2024springgrasp, chen2024bodex} that develops custom objective functions and optimizers to refine the hand pose, we propose a novel method based on the transposed Jacobian control in MuJoCo. This approach is key to achieving rich contacts while ensuring no penetration, with minimal coding effort and parameter tuning. 

Next, we evaluate the synthesized grasps in MuJoCo to assess their ability to withstand external forces applied to the object. Unlike previous work~\cite{wang2023dexgraspnet, zhang2024dexgraspnet} that designs heuristics to squeeze the hand for applying force on the object, which is not suitable for all grasp types, we design a contact-aware control strategy that computes the desired forces for each contact point and controls the hand to apply them approximately, also based on the transposed Jacobian control. Finally, high-quality grasps that pass the simulation tests can be used to construct new grasp templates, reducing the need for human annotations and broadening the range of objects that can be grasped.

Experiments show that our method greatly outperforms previous type-unaware grasp synthesis baselines in simulation, especially under more challenging testing conditions (i.e., smaller friction coefficients) and with a wider variety of objects (i.e., from Objaverse~\cite{deitke2023objaverse}). Using our proposed pipeline, we also build a large-scale dataset covering different grasp types. This dataset further enables training a type-conditional generative model that generates desired grasp types for novel objects from single-view point clouds, achieving a success rate of $82.3\%$ on the Shadow hand in real-world experiments. Finally, we show that our algorithm can be used to develop an annotation UI for collecting semantic grasps on the specified object regions with only a few mouse clicks.

In summary, our main contributions are: 

\begin{itemize} 
    \item An efficient pipeline to synthesize high-quality grasps for any grasp type, object, and articulated hand, starting from one human-annotated template per hand and grasp type. 
    \item A large-scale dataset containing 9.5M grasps and 10.7k objects, covering 31 grasp types in the GRASP taxonomy. 
    \item A type-conditional generative model that can use the specified grasp types to grasp novel objects in the real world, with only a single-view point cloud as input. 
    \item An annotation UI for collecting semantic grasps with only a few mouse clicks.
\end{itemize}

\section{Related Work}

\subsection{Analytical Grasp Synthesis}
Analytical grasp synthesis aims to find good grasps, typically measured by wrench-based force closure metrics~\cite{ferrari1992planning, roa2015grasp}. These methods generally assume complete knowledge of the object's geometry for the metric calculation, making them more suitable for data preparation than for real-world applications, where objects are often only partially observed. 
For parallel grippers and suction cups, where hand-object contact patterns are simple and the dimension of the hand pose is low (around 7), many work~\cite{mahler2016dex, mahler2018dex, fang2020graspnet, cao2021suctionnet} randomly sample a large number of grasps and select those with better metrics as the final results. 
For dexterous hands, which have more complex contact patterns and higher pose dimensions (often exceeding 20), random sampling is insufficient to find good grasps, necessitating optimization. Early work~\cite{miller2004graspit, ciocarlie2007dexterous} used sampling-based optimization, such as simulated annealing, since the commonly used metrics are non-differentiable. More recent approaches~\cite{liu2021synthesizing, turpin2022grasp, turpin2023fast, wang2023dexgraspnet, chen2023task, li2023frogger, chen2024springgrasp, chen2024bodex} introduce differentiable energies to leverage gradient-based optimization. However, these methods typically start from a rest pose, requiring many iterations to adjust the hand pose, which can be inefficient, easily lead to local minima, or produce unnatural poses. Our approach benefits from both random sampling and gradient-based optimization, greatly reducing the complexity.

\subsection{Grasp Taxonomy}
The GRASP taxonomy~\cite{feix2015grasp} categorizes 33 distinct grasp types based on human daily activities. Many previous analytical methods~\cite{chen2023task, chen2024bodex, li2023frogger, chen2024springgrasp} can only synthesize limited grasp types, such as fingertip grasp. Although some work~\cite{turpin2022grasp, turpin2023fast, liu2021synthesizing, wang2023dexgraspnet} supports contact beyond the fingertip, they can only synthesize some simple grasp types, e.g. warping fingers around the object, and typically have high randomness and cannot specify a desired type. Some work on functional dexterous grasping~\cite{jian2023affordpose, wei2024learning, agarwal2023dexterous, wu2024cross, zhu2021toward} tackles more complex, human-like grasp types, but is often limited to axis-aligned objects with similar geometries, hindering scalability. In contrast, our method can synthesize all 33 grasp types without assumptions about the object, requiring only one template per grasp type.

\subsection{Data Collection for Learning-based Grasp Synthesis}
Learning-based dexterous grasp synthesis~\cite{xu2023unidexgrasp, wu2022learning, weng2024dexdiffuser} is often built on generative models such as CVAE, diffusion models, and normalizing flows. While they can handle partial observation and be deployed in the real world, their performance is highly dependent on the quality of the training dataset. To build dexterous grasp datasets, several approaches are explored, including teleoperation~\cite{qin2023anyteleop, cheng2407open, yang2024ace}, transferring grasps from human hands~\cite{brahmbhatt2020contactpose, qin2022dexmv, liu2022hoi4d, chao2021dexycb, yang2022oakink, chen2024vividex}, and reinforcement learning (RL)~\cite{zhang2025graspxl, wan2023unidexgrasp++, christen2022d}. Our work also presents a possible approach for scaling up grasp data collection, because our method requires less human effort than teleoperation, avoids the morphological gap seen with human hands, and is often more efficient than RL.

\subsection{Physics Simulation for Robotic Grasping}

Rigid-body physics simulators, such as Isaac PhysX~\cite{makoviychuk2021isaac}, MuJoCo~\cite{todorov2012mujoco}, Bullet~\cite{coumans2015bullet}, are widely used to validate a grasp or perform RL for dexterous hands. However, prior work rarely uses simulators for local refinement of the hand pose to improve contact with the object. Most prior work~\cite{jiang2021hand, wang2023dexgraspnet, chen2024springgrasp, chen2024bodex} develops custom energy functions and optimizers to refine hand-object contact, but they often lead to centimeter-level penetrations and require careful hyperparameter tuning. In contrast, our approach utilizes MuJoCo to refine the hand pose, achieving submillimeter-level contact convergence with minimal parameter tuning in different experiment settings. This is made possible by MuJoCo’s highly optimized framework and its second-order Newton optimizer.


\section{Preliminary of Grasp Quality Metric}

In this section, we summarize a unified formulation of grasp quality metrics widely used in recent work~\cite{liu2021synthesizing, wang2023dexgraspnet, wu2022learning, chen2024bodex}. While prior work focuses on their own formulations, our summary provides a cohesive view across them. Although our pipeline does not directly optimize this metric—unlike prior analytical methods—it plays an important role in several stages of our pipeline, such as post-filtering (Sections~\ref{sec: global init} and~\ref{sec: lo}) and the contact-aware control strategy (Section~\ref{sec:sim test}).

We assume that the object $O$ is grasped by a robot hand with $m$ contact points. For each contact $i$, let $\mathbf{p}_i \in \mathbb{R}^3$ denote the contact position, $\mathbf{n}_i \in \mathbb{R}^3$ the inward-pointing surface unit normal, and $\mathbf{d}_i \in \mathbb{R}^3$ and $\mathbf{c}_i \in \mathbb{R}^3$ be two unit tangent vectors satisfying $\mathbf{n}_i = \mathbf{d}_i \times \mathbf{c}_i$. The Coulomb friction cone $\mathcal{F}_i$ and the contact Jacobian $\mathbf{J}_{o,i}$ for the object at contact $i$ are defined as follows:
\begin{align}
\label{eq: F_pcf}
\mathcal{F}_i & = \left\{\mathbf{x}_i\in\mathbb{R}^3~|~0 \leq x_{i,1} \leq 1, x_{i,2}^2+x_{i,3}^2 \leq \mu^2 x_{i,1}^2 \right\} 
\\
\label{eq: G_pcf}
\mathbf{J}_{o,i}^T & = 
\begin{bmatrix}
    \mathbf{n}_i & \mathbf{d}_i & \mathbf{c}_i \\
    \mathbf{p}_i \times \mathbf{n}_i & 
    \mathbf{p}_i \times \mathbf{d}_i & 
    \mathbf{p}_i \times \mathbf{c}_i \\
\end{bmatrix} \in \mathbb{R}^{6\times3}
\end{align}
where $\mu$ is the friction coefficient. The friction cone $\mathcal{F}_i$ represents all feasible contact forces at contact $i$, and $\mathbf{J}_{o,i}$ maps a contact force $\mathbf{x}_i$ to a wrench $\mathbf{w}_i=\mathbf{J}_{o,i}^T\mathbf{x}_{i}$.

To balance an external wrench $\mathbf{g}\in\mathbb{R}^6$ (e.g., object gravity), the optimal contact forces $\{\mathbf{f}_i\}^m_{i=1}$ are obtained by solving the following quadratic program (QP):
\begin{align}
  (\mathbf{f}_1, ..., \mathbf{f}_m) = \underset{(\mathbf{x}_1, ..., \mathbf{x}_m)}{\arg \min}~~~&\|\sum_{i=1}^m\mathbf{J}_{o,i}^T\mathbf{x}_{i} - \mathbf{g}\|^2 \label{eq: qp force} \\
    \text{s.t.}~~~&\mathbf{x}_{i} \in \mathcal{F}_i, ~~i\in\{1,...,m\} \\
    &\Sigma_{i=1}^m x_{i, 1} \geq \lambda \label{eq: normal force constraint}
\end{align}
where $\lambda$ is a hyperparameter enforcing a minimum total normal force to avoid trivial solutions. To reduce computational complexity, the friction cone $\mathcal{F}_i$ is approximated by a pyramid, converting the problem into a linearly-constrained quadratic program (LCQP) that can be efficiently solved~\cite{bishop2024relu}.

Finally, the grasp quality metric $e$ is defined as: 
\begin{equation}
e=\|\sum_{i=1}^m\mathbf{J}_{o,i}^T\mathbf{f}_{i} - \mathbf{g}\|^2 \label{eq: qp quality}
\end{equation}
where a lower $e$ indicates a more stable or robust grasp.

\begin{figure*}
    \centering
    \includegraphics[width=\linewidth]{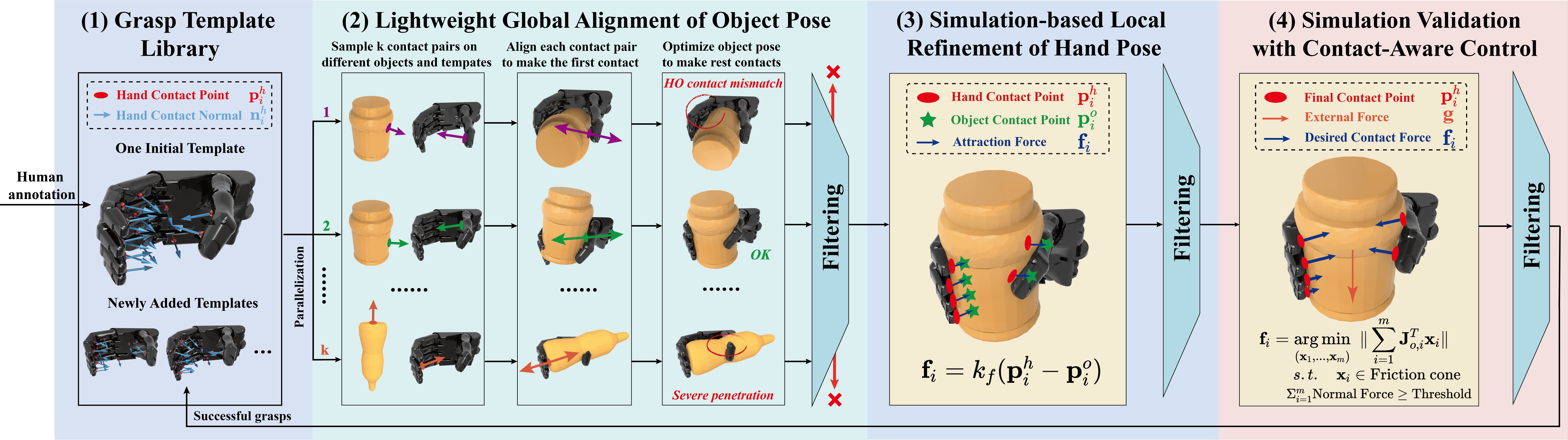}
    \caption{\textbf{The pipeline of Dexonomy.} (1) \textit{Grasp Template Library} initially requires one human-annotated template. (2) \textit{Lightweight Global Alignment} stage samples and optimizes the object poses in parallel on a GPU, to match the contact points and normals of the selected grasp templates. (3) \textit{Simulation-based Local Refinement} stage adjusts the hand pose to improve hand-object contacts. (4) \textit{Simulation Validation} tests force-closure grasps using our proposed contact-aware control strategy. (5) New templates are constructed from successful grasps and added to the \textit{Grasp Template Library}, used in the following iterations.}
    \label{fig: pipeline}
\end{figure*}

The formulation in Eq.~\ref{eq: qp force} is intuitive for testing one external wrench on the object. However, to test force-closure grasps, we typically need to consider six orthogonal gravities~\cite{chen2024bodex}. This requires solving the LCQP six times for each grasp and thus is inefficient. The energy function used in DexGraspNet~\cite{liu2021synthesizing, wang2023dexgraspnet} assumes equal contact forces and zero friction, simplifying the problem by setting $\mathbf{f}_i = [1, 0, 0]$ and $\mathbf{g} = \mathbf{0}$. While this approach eliminates the need for optimization and is extremely fast, it sacrifices accuracy. In contrast, \cite{wu2022learning} only sets $\mathbf{g} = \mathbf{0}$, providing a better balance between computational cost and accuracy. Therefore, in this paper, we adopt the variation proposed by~\cite{wu2022learning} as the grasp quality metric.

\section{Grasp Synthesis Method}

In this section, we present our proposed dexterous grasp synthesis pipeline, with an overview shown in Figure~\ref{fig: pipeline}.

\subsection{Grasp Template Definition}

A grasp template consists of several components: the hand joint configuration $\mathbf{q} \in \mathbb{R}^q$, hand contact points $\mathbf{p}^h_i \in \mathbb{R}^{3}$, corresponding normals $\mathbf{n}^h_i \in \mathbb{R}^{3}$, and the link name for each contact point ($i=1,2,\ldots,m$). Our algorithm requires a single human-annotated grasp template for each hand and grasp type as initialization. 

\subsection{Lightweight Global Alignment of Object Pose}
\label{sec: global init}

In this stage, we simultaneously sample and optimize the object pose to align with the selected template's hand contacts while keeping the hand pose fixed. The optimization variable is the object's transformation, parameterized by its (optional) scale $s_o \in \mathbb{R}$, rotation $\mathbf{R}_o \in \mathcal{S}^3$, and translation $\mathbf{t}_o \in \mathbb{R}^3$.

Before optimization, we begin with dense sampling. First, a random grasp template is selected from the \textit{Grasp Template Library}, and a random hand contact from the template is chosen. Then, a random object is selected, and a random surface point on the object is chosen. The object is initialized by aligning the sampled hand and object contacts, where contact points are matched and the contact normal directions are set opposite. The object's scale and in-plane rotation perpendicular to the normal direction are randomly sampled. Our pipeline supports parallelizing massive samples of different contacts, objects, and grasp templates on a single GPU.

During each optimization iteration, each hand contact point $\mathbf{p}^{h}_i$ calculates the nearest point $\mathbf{p}^{o}_i$ on the object's surface using the differentiable library \texttt{Warp}~\cite{warp2022}. To penalize the mismatch between hand and object contacts, we optimize the object pose by minimizing the following energy function: 
\begin{equation}
\label{eq:csample loss}
    L = k_p \sum_{i=1}^m \|\mathbf{p}^{h}_i - \mathbf{p}^o_i\|^2 + k_n \sum_{i=1}^m \|\mathbf{n}^h_i - \mathbf{n}^o_i\|^2
\end{equation}
where $k_p$ and $k_n$ are hyperparameters. There is no other energy used for optimization except Eq.~\ref{eq:csample loss}.

After optimization, results are filtered using four criteria. First, the final energy function must be below a threshold to ensure a good match between hand and object contacts. Second, severe penetration between the hand and object should be avoided, which we efficiently detect using our proposed hand collision skeletons parameterized by line segments (details in Appendix~\ref{app: post filtering}). Third, the object contact quality, as measured by Eq.~\ref{eq: qp quality}, must exceed a threshold. Finally, we apply a process similar to farthest point sampling to filter out duplicate object transformations, further detailed in Appendix~\ref{app: post filtering}.

Our design, using only one energy during optimization and leaving other checks for post-filtering, provides several advantages. First, it reduces computational costs, enabling maximized parallelization to benefit from dense sampling to avoid local optimum traps. Second, it reduces sensitivity to hyperparameters, as post-filtering criteria are applied sequentially, while optimization energies need to be applied together.

\begin{table*}[]
    \centering
    \begin{tabular}{l|c|c|c|c|c|c|c|c}
       Dataset & Hand & Sim./Real  & Objects & Grasps & Grasp Types & Force Closure & Data Type & Method \\
        \hline
        DexGraspNet~\cite{wang2023dexgraspnet} & Shadow & IsaacGym &5.4k & 1.32M & Random & \checkmark & Grasp pose & Optimization\\
        RealDex~\cite{liu2024realdex} & Shadow & Real & 52 & 59k & Random & \ding{55} & Motion & Teleoperation \\
        GraspXL~\cite{zhang2025graspxl} & Multiple & RaiSim & 500k & 10M & Random & \ding{55} & Motion & RL \\
        BODex~\cite{chen2024bodex} & Shadow & MuJoCo & 2.4k & 3.62M & Fingertip & \checkmark & Pre-grasp, grasp poses & Optimization\\

         \hline 
        Dexonomy (Ours) & Shadow & MuJoCo & 10.7k & 9.5M & 31 types & \checkmark & Pre-grasp, grasp, squeeze poses & Sampling+opt.
    \end{tabular}
    \caption{\textbf{Dexterous Grasp Dataset Comparison.} Our large-scale dataset aims to support the study of data-driven methods for type-aware grasp synthesis.}
    \label{tab: dataset statistics}

\end{table*}

\subsection{Simulation-based Local Refinement of Hand Pose}
\label{sec: lo}

In this stage, the object is fixed, and the hand pose is locally refined to improve the hand-object contact. A virtual force $\mathbf{f}_i$ is needed at each hand point $\mathbf{p}_i^h$ toward the corresponding nearest object point $\mathbf{p}_i^o$. To apply these virtual forces in MuJoCo, they are transferred to the hand's joint torque via simplified transposed Jacobian control: \begin{equation} 
\mathbf{f}_i = k_f(\mathbf{p}^h_i - \mathbf{p}^o_i), ~~~ \mathbf{\tau} = \sum_{i=1}^m \mathbf{J}_{h,i}^T \mathbf{f}_i 
\label{eq: jacb} 
\end{equation} 
where $k_f$ is a hyperparameter and $\mathbf{J}_{h,i}^T \in \mathbb{R}^{q\times 3}$ is the transpose of the hand contact Jacobian that maps force vectors from the world to joint coordinates.

While Eq.~\ref{eq: jacb} is a simplified control strategy with many assumptions (e.g., no dynamics or gravity; joint torques mapped from each contact force are independent and additive), it serves our need for synthesizing contact-rich grasps in simulation. This is easy to implement and theoretically works for other physics simulators. Eq.~\ref{eq: jacb} is iteratively applied for 200 steps, where $\mathbf{p}_i^o$ is fixed in the world frame to avoid drift, while $\mathbf{p}_i^h$ is fixed in the hand link frame (and moves in the world frame).

To ensure the hand remains strictly penetration-free with respect to the object, we apply a $1\text{mm}$ contact margin in MuJoCo—meaning the hand is repelled if it comes within $1\text{mm}$ of the object surface. Since MuJoCo reliably resolves penetration at the millimeter scale, the resulting contact distance between the hand and the object typically falls within the range of $[0, 2\text{mm}]$. This behavior motivates our use of a $2\text{mm}$ contact tolerance: a hand link is considered to be in contact if its distance to the object is less than $2\text{mm}$.

After optimization, we filter the results based on three criteria. First, there should be no penetration between the hand and the object, measured using collision meshes. Second, those fingers with at least one annotated contact are required to be in contact with the object. This prevents the synthesized grasps from deviating significantly from the initial template. Finally, the grasp quality, as measured by Eq.~\ref{eq: qp quality}, must exceed a threshold, which is needed again because the final contact points may slightly differ from the ones in the previous stage.

\subsection{Simulation Validation with Contact-Aware Control}
\label{sec:sim test}
To validate the synthesized grasps in MuJoCo, the hand should squeeze to hold the object stably, controlled by a control signal of joint torques. Our contact-aware control strategy first calculates the desired forces on each contact using Eq.~\ref{eq: qp force} where $\mathbf{g}=\mathbf{0}$, and then converts these forces into joint torques using the same transposed Jacobian control as in Eq.~\ref{eq: jacb}. 

The $\mathbf{g}$ is set to zero to reduce computational cost and standardize the control signals across different external forces. Despite the resultant force of all contacts being approximately zero, friction naturally adjusts to prevent object movement. Given the large strength of the normal force (i.e., around $1\sim10$N in our experiment for a 100g object), constrained by Eq.~\ref{eq: normal force constraint}, there is typically sufficient room for friction to adjust, if the grasp pose is physically plausible. A grasp is considered to succeed only if the object remains stable under all six orthogonal external forces for 2 seconds in simulation.

\subsection{Construction of New Grasp Templates}

Once a grasp successfully passes the simulation validation, a new grasp template is constructed and added to the template library. The joint configuration of the new template is taken from the successful grasp, while the contact information (i.e., points and normals) is updated only if an actual contact is detected near the original contact on the same hand link. This strategy prevents the new template's contact information from deviating too much from the original. Newly added templates can be randomly selected during the global alignment stage in subsequent iterations of the whole pipeline.

\begin{figure}
    \centering
    \includegraphics[width=1.0\linewidth]{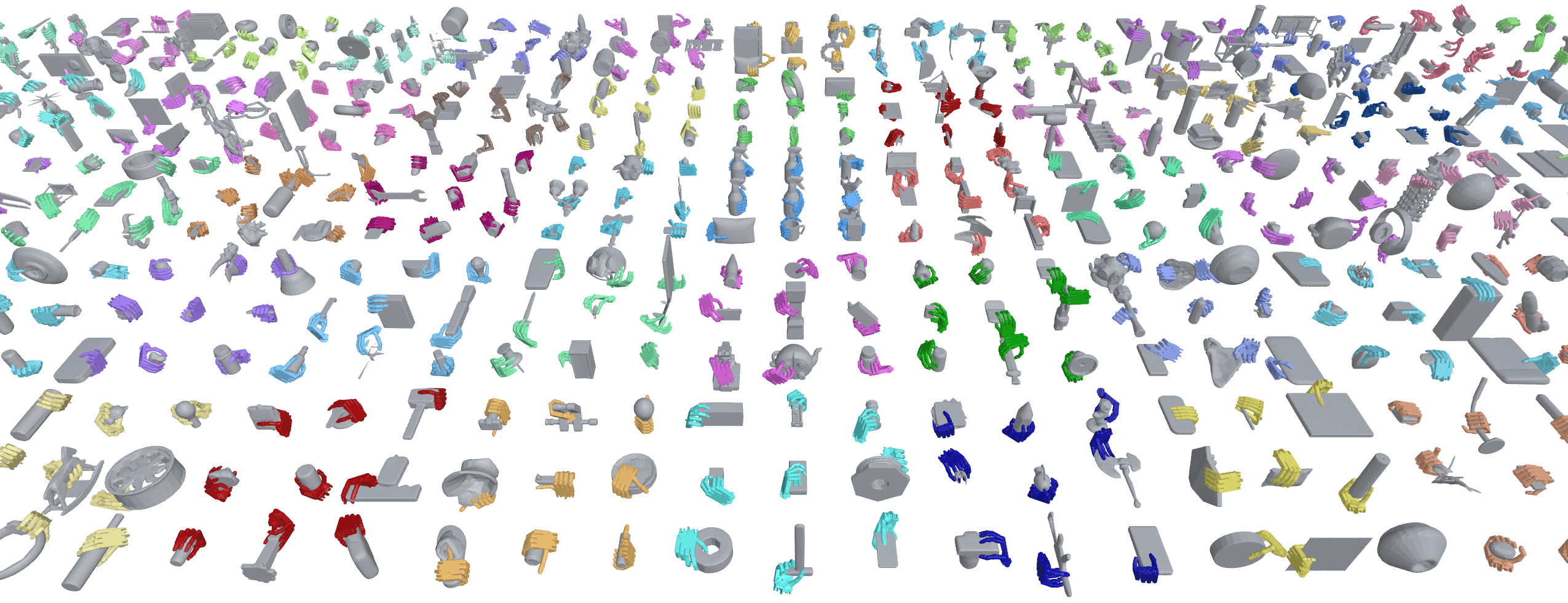}
    \caption{\textbf{Dexonomy Dataset Visualization.} Each color corresponds to a different grasp type.}
    \label{fig:dexonomy_dataset}
\end{figure}

\section{Dexonomy Dataset}
\label{sec: dexonomy dataset}

Using our proposed grasp synthesis pipeline, we construct a large-scale dataset for Shadow hand covering 31 grasp types from the GRASP taxonomy~\cite{feix2015grasp}. This dataset is designed to support research on data-driven methods for type-aware grasp synthesis. Two grasp types in the taxonomy, \textit{Distal Type ($\#19$)} and \textit{Tripod Variation ($\#21$)}, are excluded due to their specificity to object categories, namely scissors and chopsticks, respectively.

As shown in Table~\ref{tab: dataset statistics}, our dataset comprises 10.7k object assets, including 5,697 objects from DexGraspNet~\cite{wang2023dexgraspnet} and 5,000 new objects randomly selected from Objaverse~\cite{deitke2023objaverse}. All objects are normalized such that the diagonal of their axis-aligned bounding box is 2 meters, with scales ranging between $[0.05, 0.2]$. Only successful grasps are retained, resulting in 9.5M data points. The entire dataset was synthesized in less than 3 days on a server with 8 NVIDIA RTX 3090 GPUs. Additional statistics are provided in Table~\ref{tab: grasp taxonomy statistic} and Appendix~\ref{app: dexonomy dataset details}.

Each data point includes three key poses:
\begin{itemize}
    \item \textbf{Grasp pose}, obtained via local refinement (Section~\ref{sec: lo}).
    \item \textbf{Pre-grasp pose} for collision-free motion planning, generated after the grasp pose by enforcing a $2\text{cm}$ contact margin in MuJoCo—pushing the hand away if it is within $2\text{cm}$ of the object.
    \item \textbf{Squeeze pose}, derived from the control signal used for simulation validation (Section~\ref{sec:sim test}), to apply force through hand-object contacts.
\end{itemize}
These poses provide the minimal requirements for generating a complete grasping trajectory (including reaching and squeezing) and are compatible with diverse robot arms and initial hand configurations.

\begin{figure}
    \centering
    \includegraphics[width=1.0\linewidth]{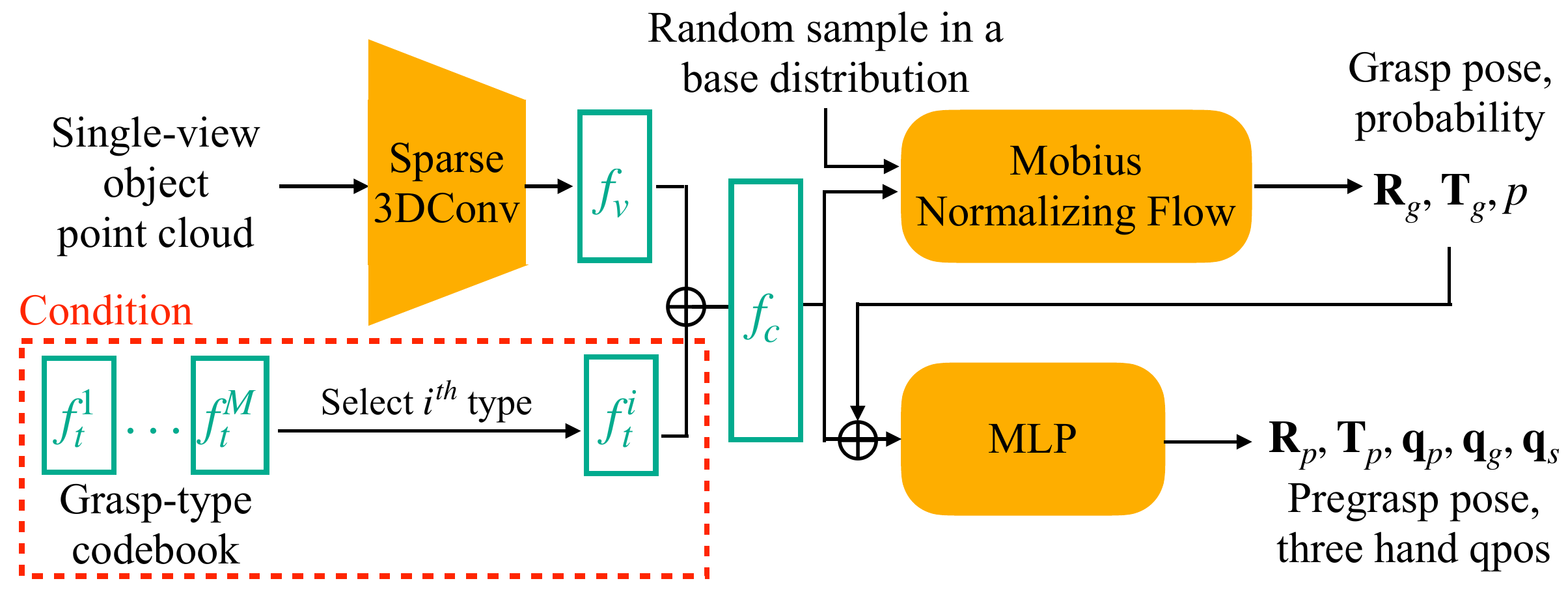}
    \caption{\textbf{Type-Conditional Grasp Generative Model.} Without the grasp-type codebook in the red dashed box, the model becomes type-unconditional and is similar to BODex~\cite{chen2024bodex}. }
    \label{fig: learning model}
\end{figure}

\begin{table*}[]
    \centering
    \begin{tabular}{l|c|c|c|c|c|c|c|c}
         & GSR ($\%$) $\uparrow$ & OSR ($\%$) $\uparrow$ & S ($s^{-1}$) $\uparrow$ & CLN $\uparrow$ & CDC ($mm$) $\downarrow$ & PD ($mm$) $\downarrow$ & SPD ($mm$) $\downarrow$ & D ($\%$) $\downarrow$ \\
         \hline
        DexGraspNet~\cite{wang2023dexgraspnet} 
&12.10
&57.01
&3.25
&3.22
&7.58
&4.85
&1.20
&29.03
\\
        FRoGGeR~\cite{li2023frogger} 
&10.34
&55.70
&2.98
&2.51
&4.95
&0.22
&\textbf{0.00}
&\textbf{27.01}
\\
        SpringGrasp~\cite{chen2024springgrasp} & 7.83
&35.44
&5.47
&2.79
&23.59
&16.58
&1.06
&70.18\\
        BODex~\cite{chen2024bodex} & 49.23
&\textbf{96.56}
&\textbf{403.9}
&3.85
&3.03
&0.63
&0.02
&32.50\\
        \hline
        Ours & \textbf{60.50}
& 96.53
& 323.4
& \textbf{4.38}
& \textbf{0.21}
&\textbf{0.00}
&\textbf{0.00}
& 34.17
    \end{tabular}
    \caption{\textbf{Comparison with Type-Unaware Grasp Synthesis Baselines for Allegro Hand.} Most baselines, except DexGraspNet, only synthesize fingertip grasps, so we also synthesize fingertip grasps for a fair comparison.}
    \label{tab: fingertip baseline}
\end{table*}

\section{Type-Conditional Grasp Generative Model}

To generate grasps from partial observations for real-world deployment, data-driven methods are essential. Although learning is not the main focus of this paper, we present a simple model as an initial try. The model architecture, illustrated in Figure~\ref{fig: learning model}, is similar to previous work~\cite{zhang2024dexgraspnet, chen2024bodex}, with the key difference being the grasp-type codebook added as a conditional input to specify a grasp type. 

The input to the model consists of a single-view object point cloud and a type feature $f_t^i$ selected from the grasp-type codebook. The point cloud is encoded into a feature $f_v$ using a Sparse3DConv network with MinkowskiEngine~\cite{choy20194d}. This vision feature $f_v$, along with the type feature $f_t$, are concatenated to form a conditional feature $f_c$. Conditioned on $f_c$, the Mobius normalizing flow~\cite{liu2023delving} maps a random sample in a base distribution to a grasp pose $\mathbf{R}_g$ and $\mathbf{T}_g$, and calculates a probability $p$ indicating the pose quality. The predicted grasp pose is then concatenated with $f_c$ and passed through an MLP to predict a pre-grasp pose $\mathbf{R}_p$, $\mathbf{T}_p$, and three hand qpos $\mathbf{q}_p$, $\mathbf{q}_g$, and $\mathbf{q}_s$ for the pre-grasp, grasp, and squeeze poses, respectively. The whole model is trained end-to-end and the type feature $f_t^i$ is also optimizable.

\section{Experiment}

\subsection{Evaluation Metrics}
\label{sec: metrics}
The following metrics are used for a comprehensive evaluation of the synthesis pipeline and grasp quality. All distances are measured using collision meshes in MuJoCo.

\textbf{Grasp Success Rate} (GSR) (unit: $\%$): The percentage of successful grasps relative to the attempt number. For our method, one attempt is defined as one valid result output by the global alignment stage. A grasp succeeds only if it resists six external forces in MuJoCo and does not have severe penetrations ($>1$ cm), since the penetration may cause simulation failure and prevent the object from moving. The object mass is 100g, and the success criteria for the object pose are $5\text{cm}$ and $15^\circ$.

\textbf{Object Success Rate} (OSR) (unit: $\%$): The percentage of objects that have at least one successful grasp. If the object scales are fixed, as in Sections~\ref{sec: type unaware exp} and~\ref{sec: learning exp}, different scales of the same object are treated as separate objects.

\textbf{Speed} (S) (unit: second$^{-1}$): The maximum number of attempts completed per second on a server with 8 NVIDIA RTX 3090 GPUs and 2 Intel Xeon Platinum 8255C CPUs (48 cores, 96 threads). We report the time running on a server because our method utilizes both GPUs and CPUs. This metric excludes simulation validation.

\textbf{Contact Link Number} (CLN): The number of hand links whose distance to the object surface is within 2 mm.

\textbf{Contact Distance Consistency} (CDC) (unit: mm): The delta between the maximum and minimum signed distances across all fingers. This metric quantifies the variation in contact distance across different fingers and is invariant to penetration.

\textbf{Penetration Depth} (PD) (unit: mm): The maximum intersection distance between the hand and object for each grasp.

\textbf{Self-Penetration Depth} (SPD) (unit: mm): The maximum self-intersection distance among different hand links.

\textbf{Diversity} (D) (unit: $\%$): The proportion of total variance explained by the first principal component in PCA, computed as the ratio of the first eigenvalue to the sum of all eigenvalues. PCA is performed on data points that include grasp translation $\mathbf{T}_g$, rotation $\mathbf{R}_g$ (in the axis-angle representation), and joint angles $\mathbf{q}_g$.

\subsection{Type-Unaware Grasp Synthesis}
\label{sec: type unaware exp}

\subsubsection{Comparison with analytical methods} Four open-source baselines are compared: DexGraspNet~\cite{wang2023dexgraspnet}, FRoGGeR~\cite{li2023frogger},  SpringGrasp~\cite{chen2024springgrasp}, and BODex~\cite{chen2024bodex}. 

\textbf{Experiment settings.} Most of the settings follow the benchmark provided by BODex~\cite{chen2024bodex}, with some modifications on the object set to increase difficulty. The Allegro hand is used, as it is the only hand type supported by all baselines. 5697 object assets from DexGraspNet are used, with six scales applied to each normalized object: $0.05$, $0.08$, $0.11$, $0.14$, $0.17$, and $0.20$. Although our method supports optimizing object scales, we fix the object scale to ensure a fair comparison. The attempt number for each method is set to 20, and we manually annotate two grasp templates, each with 10 attempts.

\textbf{Quantitative result analysis.} As shown in Table~\ref{tab: fingertip baseline}, our method achieves the highest grasp success rate and best performance on contact and penetration. Our speed is slightly lower than that of BODex, as their pipeline is highly optimized for GPUs, while our local refinement stage uses MuJoCo's CPU version. Detailed time analysis of our pipeline is available in Appendix~\ref{app: time analysis}. Our diversity is somewhat lower, as we use only two initial templates and our hand refinement stage makes only local adjustments. However, the overall diversity of our Dexonomy dataset is much better, as reported in Table~\ref{tab: grasp taxonomy statistic}. The success rates of baselines are lower than those reported in BODex~\cite{chen2024bodex}, primarily because our objects have a higher mass (100g vs. 30g), a larger scale range ($[0.05, 0.2]$ vs. $[0.06, 0.12]$), and more diverse shapes (5697 instances vs. 2397). More details about the performance drop of baselines are provided in Appendix~\ref{app: bodex performance drop}.

\begin{figure}[t]
    \centering
    \includegraphics[width=\linewidth]{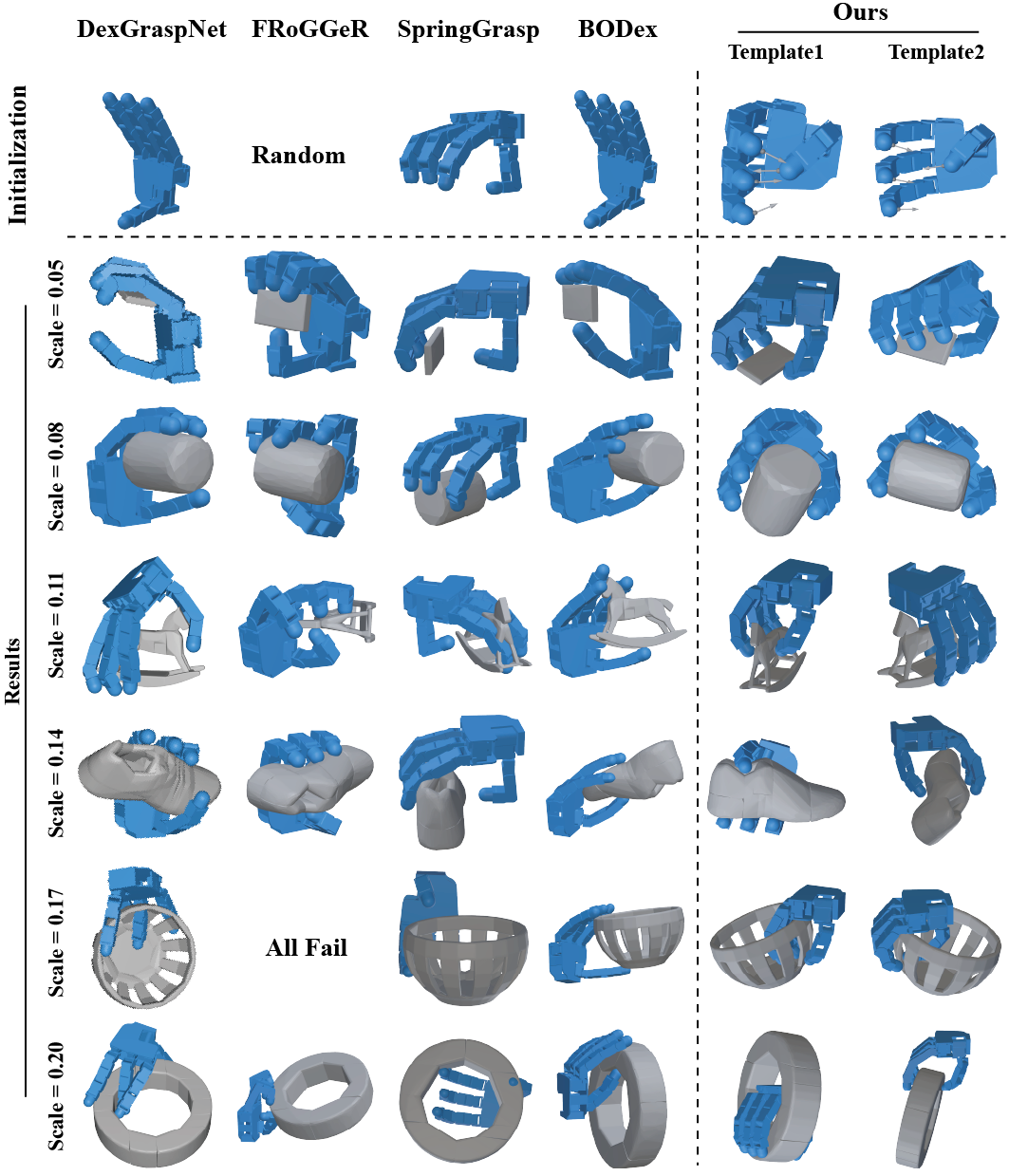}
    \caption{\textbf{Visualization of Type-Unaware Grasps.} Our method synthesizes human-like and stable grasps, even for objects with complex geometries (e.g., object scales = 0.11, 0.17, and 0.20).}
    \label{fig: vis common}
\end{figure}

\textbf{Visualization analysis.} Figure~\ref{fig: vis common} illustrates the initial hand pose and some synthesized grasps for each method. Our method consistently synthesizes human-like and stable grasps, even for objects with complex geometries (e.g., for scales 0.11, 0.17, and 0.20). Notably, the synthesized grasp for template 1 and object scale 0.05 requires high precision and is challenging for previous methods. Furthermore, the grasp for template 2 and object scale 0.14 shows a much larger thumb-to-other-tip distance than the initial human-annotated template, demonstrating our method's ability to adjust hand joint angles across a large range.

For baseline methods, DexGraspNet~\cite{wang2023dexgraspnet} shows high uncertainty, partly due to its randomness in selecting contact points. While it occasionally generates good grasps (e.g., for scales 0.08 and 0.14), it often results in twisted fingers (e.g., for scales 0.17 and 0.20) or large thumb-to-object distance (e.g., for scale 0.05). FRoGGeR~\cite{li2023frogger} performs well on simple objects but almost always fails on objects with complex geometries. It also tends to generate grasps with different contact normals for each fingertip (e.g., for scales 0.05 and 0.08), an issue encouraged by many previous force closure metrics. SpringGrasp~\cite{chen2024springgrasp} suffers from severe penetration and inconsistent contact distances, especially for the thumb. Additionally, their grasps lack diversity, and their thumb joint frequently exceeds the feasible range, which is not executable in both MuJoCo and the real world. Although BODex~\cite{chen2024bodex} demonstrates high success rates in simulation, their synthesized grasps rarely involve finger bending, resulting in unnatural poses.

\begin{table}[]
    \centering
    \begin{tabular}{l|c||c|c|c|c}
        \multirow{2}{*}{Method} & Attempt & \multicolumn{2}{c|}{DGN object~\cite{wang2023dexgraspnet}} & \multicolumn{2}{c}{Objaverse~\cite{deitke2023objaverse}} \\
         & Number & GSR$\uparrow$ & OSR$\uparrow$ & GSR$\uparrow$ & OSR$\uparrow$  \\
         \hline
        BODex & 20 &14.79
& 71.30&6.92
&43.48
\\
        BODex & 100 &14.80
& 89.84&6.91
&73.53
\\
\hline
        Ours & 20 &27.16
& 91.28&18.25 
&84.17
\\
        Ours & 100 &\textbf{27.18}
& \textbf{95.13}&\textbf{18.34}
&\textbf{94.63}
\\

    \end{tabular}
    \caption{\textbf{A Harder Benchmark for Fingertip Grasp Synthesis.} This benchmark uses smaller friction coefficients and more diverse objects, and our method consistently outperforms the baseline. DGN indicates DexGraspNet.}
    \label{tab: fingertip objaverse}

\end{table}

\subsubsection{Comparison using a harder benchmark} The benchmark in Table~\ref{tab: fingertip baseline} uses large friction coefficients and many simple objects, which do not fully reflect the ability of each method to synthesize very high-quality grasps in more complex scenarios. To address this, we introduce a more challenging benchmark by reducing the tangential and torsional friction coefficients from 0.6 and 0.02 to 0.3 and 0.002, respectively, and randomly selecting 5000 additional objects from Objaverse~\cite{deitke2023objaverse} for testing. To mitigate the increased difficulty, we allow each method more attempts per object (from 20 to 100). We compare only with BODex, as other baselines exhibit significantly lower success rates and slower speeds.

As shown in Table~\ref{tab: fingertip objaverse}, our method significantly outperforms BODex. Notably, our method achieves an object success rate exceeding $94\%$, successfully grasping nearly all scaled objects, while BODex fails on about $27\%$ of the Objaverse objects. This highlights our method's stronger generalizability to complex in-the-wild objects. Additionally, our grasp success rate continues to improve with more attempts, benefiting from continuous updates to our template repository, whereas the performance of BODex remains unchanged. This further demonstrates the adaptability of our approach.

\subsection{Type-Aware Grasp Synthesis}

\begin{figure}[t]
    \centering
    \includegraphics[width=1.0\linewidth]{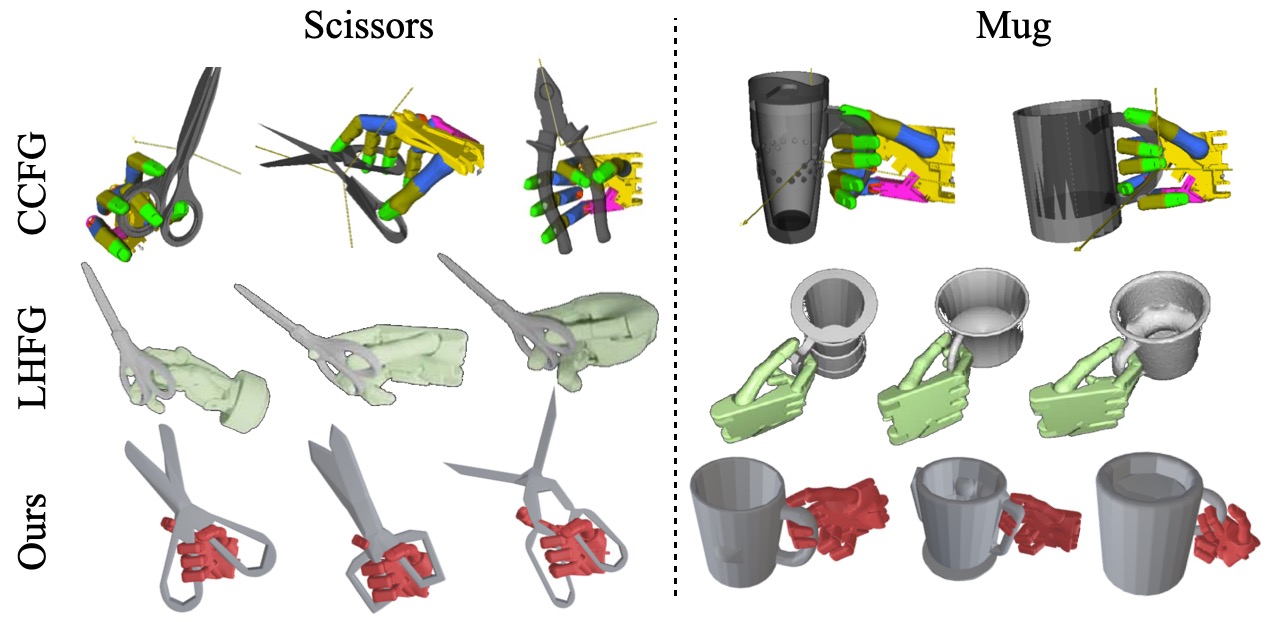}
    \caption{\textbf{Comparison with Functional Grasp Transfer Baselines.} Our grasps involve more contact points without any penetration, indicating higher stability, especially for scissors.}
    \label{fig: functional comparison}
\end{figure}

\subsubsection{Comparison with functional grasp transfer baselines} 

Unfortunately, we did not acquire the code of suitable baselines for comparison, as existing methods either do not support robotic hands (e.g., Oakink~\cite{yang2022oakink}) or have not made their code publicly available (e.g., LHFG~\cite{wei2024learning} and CCFG~\cite{wu2024cross}). Consequently, we can only perform qualitative comparisons using figures from their papers. As shown in Figure~\ref{fig: functional comparison}, previous baselines mainly use fingertips to grasp the object, particularly for scissors. In contrast, our method achieves significantly more contact points (approximately 10 links in contact for scissors and 7 for mugs), resulting in more stable and human-like grasps. Additionally, CCFG’s grasps show noticeable penetrations, especially with mugs, while LHFG reports a maximum penetration of around 1 cm in their paper. In contrast, our grasps do not have any penetration.

\begin{table}[]
    \centering
    \begin{tabular}{c|c|c|c|c|c|c}
         &  \multicolumn{2}{c|}{GSR($\%$)$\uparrow$} & \multicolumn{2}{c|}{OSR($\%$)$\uparrow$} & \multirow{2}{*}{CLN$\uparrow$} & \multirow{2}{*}{D($\%$)$\downarrow$} \\
         &  Normal & Hard & Normal & Hard &  &  \\	
         \hline 
        Power & 24.2 & \textbf{12.8} & 81.9 & 68.3 & \textbf{9.1} & \textbf{24.7}	\\
        Intermediate & 23.0 & 6.6	& 79.9&69.4 & 4.8 & 27.6 \\
        Precision & \textbf{36.0} & 11.4 & \textbf{95.9}& \textbf{85.6} & 4.2 & 25.8\\

    \end{tabular}
    \caption{\textbf{Statistics of Grasp Synthesis for the  GRASP Taxonomy.} The success rate is lower than fingertip grasps because many flexible grasp types are suitable only for specific objects, e.g., \textit{Lateral ($\#16$)} grasps for flat objects.}
    \label{tab: grasp taxonomy statistic}

\end{table}

\subsubsection{Statistics analysis of our pipeline}

In the absence of a suitable baseline for comparison, we provide some quantitative results in Table~\ref{tab: grasp taxonomy statistic}, which were gathered while synthesizing our Dexonomy dataset in Section~\ref{sec: dexonomy dataset}. The grasp types are categorized into three large groups, namely power, intermediate, and precision grasps, according to the GRASP taxonomy~\cite{feix2015grasp}.

The overall success rate is considerably lower than that of fingertip grasp synthesis, as many flexible grasp types are designed for specific object shapes. For instance, the \textit{Lateral ($\#16$)} grasp is only used for flat and small objects. Among different grasp types, precision grasps exhibit the highest success rate under \textit{normal} test conditions (i.e., with friction coefficients of 0.6 and 0.02), since these grasps typically involve only the fingertips and suit more objects. However, the success rate of precision grasps drops more rapidly than that of power grasps when the friction coefficients are reduced to 0.3 and 0.002 (i.e., the \textit{hard} test conditions), indicating that power grasps offer higher stability due to more contact with the object. Additionally, the overall diversity of grasps is better than previous work reported in Table~\ref{tab: fingertip baseline}, owing to the inclusion of many distinct grasp types.

\begin{table}[t]
    \centering
    \begin{tabular}{cc|c|c|c|c}
         & & GSR$\uparrow$ & OSR$\uparrow$ & CDC$\downarrow$ & PD$\downarrow$ \\
         \hline
       \multirow{2}{*}{Global stage} & w/o opt. & 45.7 & 88.4 & 0.22 & \textbf{0.00} \\
        & w/o filter & 18.6 & 80.3 & 0.34 & \textbf{0.00} \\
        \hline 
       \multirow{2}{*}{Local stage}  & w/o opt.& 17.8 & 72.6 & 10.24 & 4.95\\
         & w/o filter& \textbf{62.8} & \textbf{96.7} & 0.82 & 0.34 \\
        \hline 
       Template library & w/o update& 41.5 & 88.8 & 0.24 & \textbf{0.00}\\
        \hline 
       \multicolumn{2}{c|}{Ours} & 60.5 & 96.5 & \textbf{0.21} & \textbf{0.00}\\
    \end{tabular}
    \caption{\textbf{Ablation Study on Pipeline Modules for Fingertip Grasp Synthesis of Allegro Hand.}}
    \label{tab:modular ablation}
\end{table}

\subsection{Ablation Study}

\subsubsection{Modular ablation}

Table~\ref{tab:modular ablation} shows the impact of each module on fingertip grasp synthesis for the Allegro hand. For the global alignment stage, optimization provides a slight improvement, while post-filtering plays a more significant role. We also take a deeper look at the post-filtering of this stage, and find that a significant portion ($>50$k out of $80$k) is filtered out due to a high loss (Eq.~\ref{eq:csample loss}). In contrast, collision and grasp quality filters only remove fewer than 10k grasps, indicating that using them as new losses for optimization is not effective. In the local refinement stage, optimization is crucial for improving grasp quality, while post-filtering primarily ensures penetration-free grasps and negatively affects the success rate. Additionally, constructing new grasp templates significantly enhances the synthesis success rate.

\begin{table}[t]
    \centering
    \begin{tabular}{c|c|c|c|c}
      \multirow{2}{*}{GSR ($\%$) $\uparrow$}   & \multicolumn{3}{c|}{Shadow} & Allegro \\ 
       & Power & Intermediate & Precision & Fingertip \\
\hline 
        Grasp only
& 17.64		
& 9.81
& 13.79
& 24.60
\\
w/ pre-grasp
& \textbf{25.85}
& 19.93
& 34.56
& 45.08
\\
\hline
Ours & 24.23
& \textbf{23.03}
& \textbf{36.00}
& \textbf{60.50}\\
    \end{tabular}
    \caption{\textbf{Ablation on Control Strategy for Simulation Validation}. Power grasps are robust likely due to rich contacts.}
    \label{tab: abla control}
\end{table}

\subsubsection{Control strategy ablation}
In Table~\ref{tab: abla control}, we evaluate the impact of different control strategies during simulation testing. The \textit{Grasp only} method computes the pre-grasp and squeeze joint angles by scaling the grasp joint angles with fixed factors: $0.9\times$ for the pre-grasp and $1.1\times$ for the squeeze pose. The \textit{w/ pre-grasp} method calculates the squeeze joint angles as $squeeze=2\times grasp - pregrasp$, following BODex~\cite{chen2024bodex}. Note that these two strategies only calculate the joint angles, while the 6DoF hand root pose is the same as the one of the grasp pose. During execution, the hand transitions sequentially from the pre-grasp to the grasp pose, and then to the squeeze pose.

The results demonstrate that our contact-aware control strategy consistently improves the grasp success rate, except the power grasps. Power grasps are particularly robust to the control strategy, likely due to the rich contacts.

\subsubsection{Robustness to initial human-annotated templates}

\begin{figure}
    \centering
    \includegraphics[width=\linewidth]{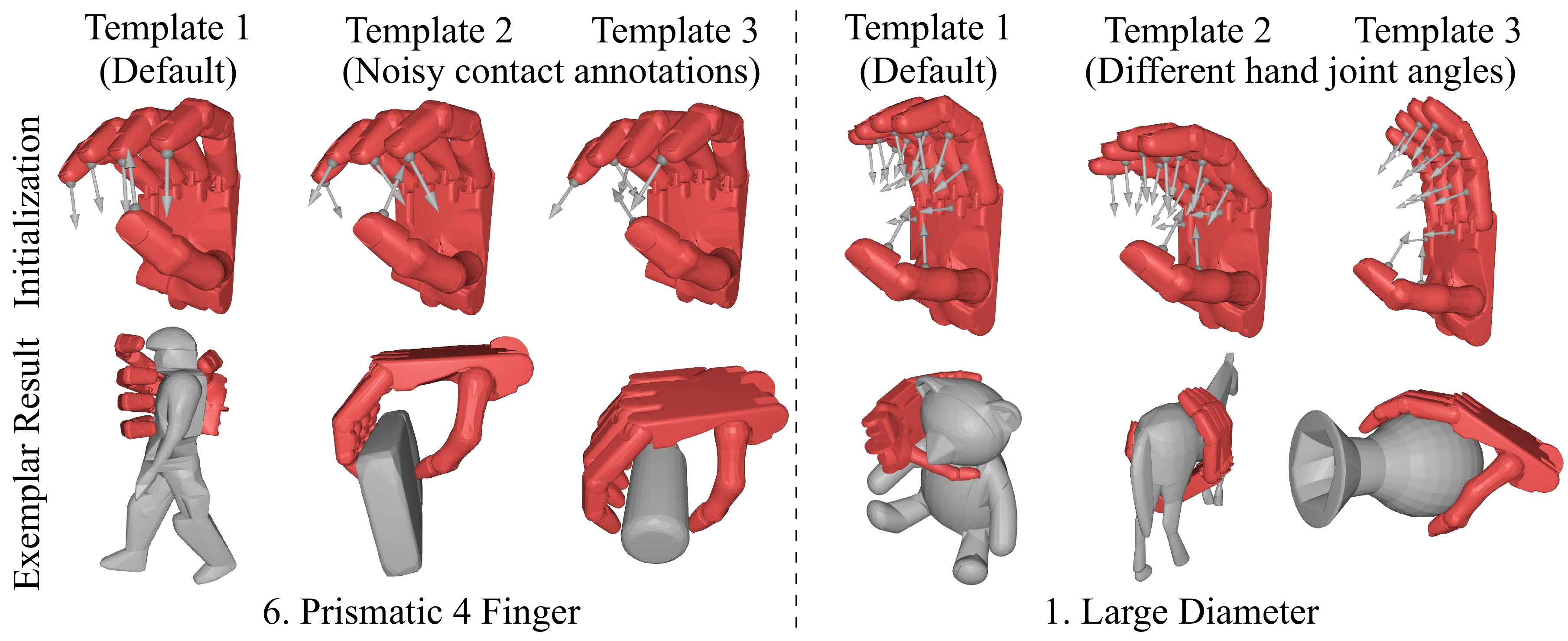}
    \caption{\textbf{Robustness to Initial Templates}. Our method generates good grasps even for very noisy contact annotations as in columns 2 and 3.}
    \label{fig: robustness to template}
\end{figure}

To investigate the robustness of our pipeline to different initial templates, we annotated two additional templates for two common grasp types of the Shadow Hand and evaluated on 1000 random objects, with 100 attempts per object. As shown in Table~\ref{tab: robustness to template} and Figure~\ref{fig: robustness to template}, our algorithm is robust to both noisy contact annotations and variations in hand joint angles, thanks to our template-adding strategy. As long as the grasp success rate is not $0\%$, the template-adding strategy will construct high-quality templates from successful grasps and greatly reduce the impact of bad initial templates.

A more challenging case is using random initialization. However, this setting fails catastrophically, likely due to the high dimensionality of our template (including hand pose and contact annotations), where valid grasps are extremely sparse. This highlights the usage of human-annotated templates.

\begin{table}[]
    \centering
    \begin{tabular}{l|c|c|c|c|c|c}
        Grasp Success Rate($\%$)$\uparrow$ & \multicolumn{3}{c|}{6.Prismatic 4 Finger} & \multicolumn{3}{c}{1.Large Diameter} \\
        \hline 
        Template ID & 1 & 2 & 3 & 1 & 2 & 3 \\
        \hline 
        w/o new templates  & 45.6 & 12.0 & 1.7 & 22.9& 22.5 & 7.2 \\
        w/ new templates (Ours)  & \textbf{57.1} & \textbf{56.7} & \textbf{46.2} & \textbf{30.1} & \textbf{34.7} & \textbf{26.8} \\
    \end{tabular}
    \caption{\textbf{Robustness to Initial Templates}. Our strategy, adding new templates from successful grasps, is the key to robustness.}
    \label{tab: robustness to template}
\end{table}

\begin{table}[t]
    \centering
    \begin{tabular}{c|c|c|c|c|c|c}
     Method  & Dataset & GSR$\uparrow$ & OSR$\uparrow$& CDC$\downarrow$& PD$\downarrow$ & D$\downarrow$ \\
         \hline 
    \multirow{4}{*}{Type-uncond.}  &  DGN~\cite{wang2023dexgraspnet} & 8.32
&44.3
&20.5
&15.9
&29.1\\
   &    BODex~\cite{chen2024bodex} & 54.0
& 84.4
& 11.7
&\textbf{6.2}
& 32.0\\
 & Ours-type1 & 55.5
&85.9
&\textbf{10.8}
&8.4
&31.5\\

      & Ours-all & 24.5
&73.2
&15.6
&11.6
&28.0
\\
\hline 
   Type-cond. & Ours-all &\textbf{63.9}
&\textbf{91.3}
&13.9
&8.6
&\textbf{25.7}\\
    \end{tabular}
    \caption{\textbf{Learning-based Grasp Synthesis from Single-View Object Point Clouds in Simulation.} Our type-conditional model trained on our Dexonomy dataset significantly outperforms baselines.}
    \label{tab:learning}
\end{table}

\subsection{Learning-based Grasp Synthesis from Partial Observation}
\label{sec: learning exp}
In this section, we compare the influence of both the grasping dataset and the learning method in simulation. The 10.7k objects in our Dexonomy dataset are randomly split into training and test sets with a 4:1 ratio. While the object scales used for training vary, we fix the scales during testing, using the same six scale levels as described in Section~\ref{sec: type unaware exp}. To ensure a fair comparison, we also regenerate a dataset for BODex using our objects and scales, resulting in 0.7M valid grasps.
\textit{Ours-type1} includes only the \textit{Large Diameter ($\#1$)} grasp type from the Dexonomy dataset and contains 0.4M data points, while \textit{Ours-all} uses the full 9.5M dataset.
For the type-conditional model, we additionally train a classifier to select the best grasp type based on each object’s point cloud; more details are provided in Appendix~\ref{app: learning design}.
For each object, 100 candidate grasps are predicted and ranked by their associated probabilities, with the top 10 selected as the final outputs.

\begin{figure*}
    \centering
    \includegraphics[width=1.0\linewidth]{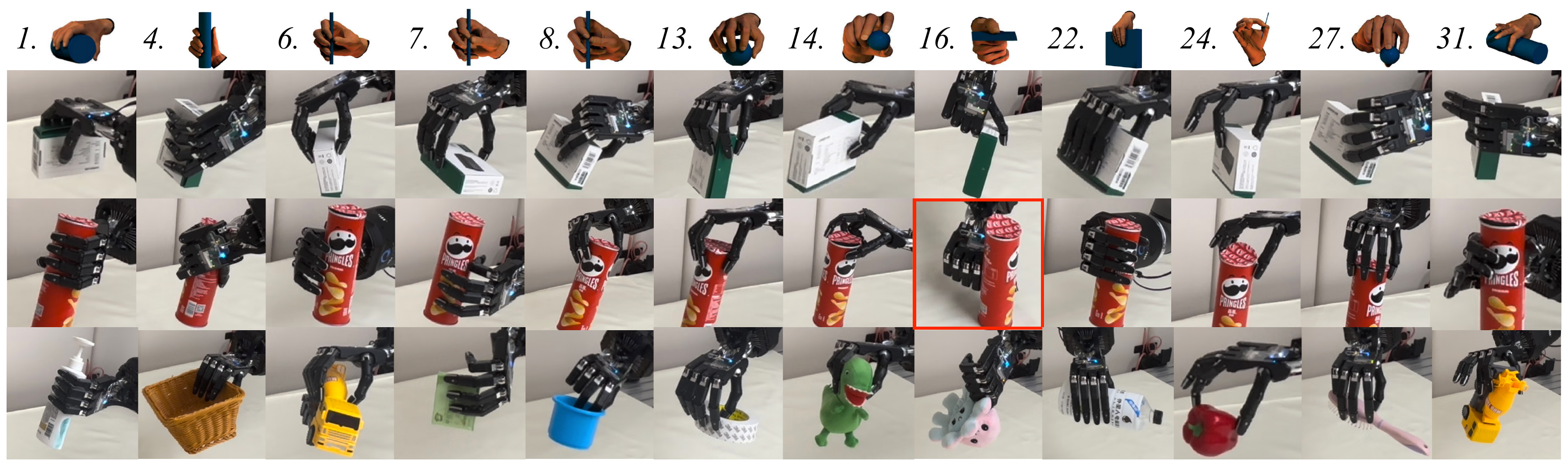}
    \caption{\textbf{Real-World Gallery.} Our trained type-conditional generative model successfully synthesizes desired grasp types from single-view object point clouds. All grasps succeed in lifting the object except the one in the red box, where the grasp type is unsuitable for the object.}
    \label{fig: real world gallery}
\end{figure*}

As shown in Table~\ref{tab:learning}, our type-conditional model trained on the Dexonomy dataset significantly outperforms the BODex baseline by around $10\%$, further highlighting the value of our dataset. Notably, even when using only a single grasp type with less data, the learned model still outperforms its counterpart trained on BODex. Without type-conditional features, the model struggles to learn from the diverse grasp data and performs poorly. In contrast, the type-conditional model successfully synthesizes the intended grasp types, as visualized in Figure~\ref{fig: real world gallery}. The model trained on our dataset exhibits slightly higher penetration, likely due to the fact that our grasps are more contact-rich. Contact distance consistency is also higher for \textit{Ours-all}, as this metric considers all fingers, while some grasp types do not involve every finger.

\begin{figure}
    \centering
    \includegraphics[width=0.95\linewidth]{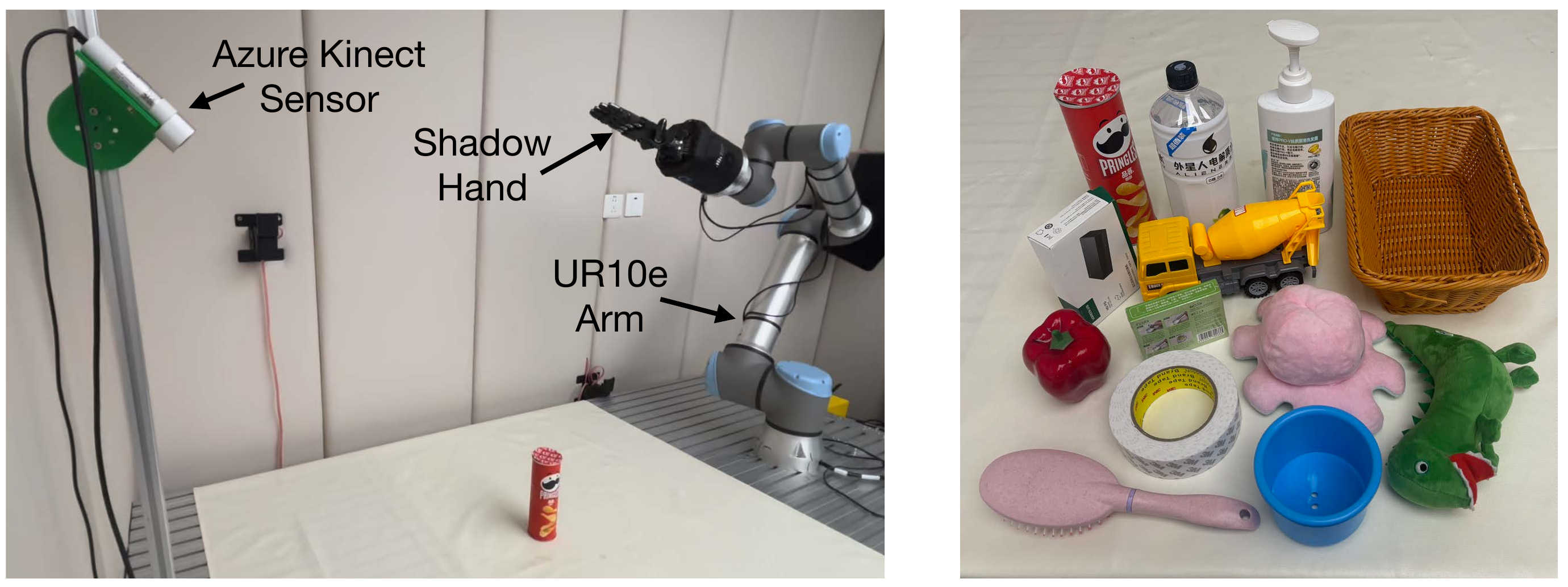}
    \caption{\textbf{Real-World Experiment.} (Left) Hardware setup. (Right) 13 varied objects for testing.}
    \label{fig: real world setup}
\end{figure}

\subsection{Real-World Experiment}

Finally, we validate the trained model in the real world. The experimental setup and 13 unseen objects for testing are shown in Figure~\ref{fig: real world setup}. To simplify the testing process, we select 12 grasp types from the taxonomy that are distinct and suitable for grasping from a table. For 3 simple objects, namely the red cylinder, the white box, and the red apple, we use all 12 grasp types to grasp them. For the other 10 objects, we let GPT-4o choose 3 suitable grasp types by providing pictures of the grasp types and the objects. Each object and grasp type combination is tested in 3 trials, resulting in around 200 total testing trials.

To perform a grasp, a single-view object point cloud segmented by SAM2~\cite{ravi2024sam} and the specified grasp type are taken as input to the trained type-conditional generative model. The model generates 100 candidates and we use the pre-grasp poses as the target for collision-free motion planning with CuRobo~\cite{sundaralingam2023curobo}, filtering out failed ones. The remaining grasps are ordered by the output probability of the normalizing flow, and the top 3 are executed. In this way, we prevent the success rate of motion planning from affecting the results, since it is not the focus of this paper. After reaching the pre-grasp pose, the hand moves to the grasp pose and then the squeeze pose to grasp the object stably, and finally lifts it.

As shown in Figure~\ref{fig: real world gallery}, our model can correctly generate physically plausible grasps for the specified types and achieves an overall success rate of $82.3\%$. 
The most common failure mode is the unsuitable grasp types for the object, especially for \textit{Lateral ($\#16$)}. Moreover, some grasp types, such as \textit{Tip Pinch ($\#24$)}, are not very robust and easily fail due to real-world noise and prediction errors. More details are provided in Appendix~\ref{app: real world details}.

\section{Application}

\begin{figure}
    \centering
    \includegraphics[width=0.9\linewidth]{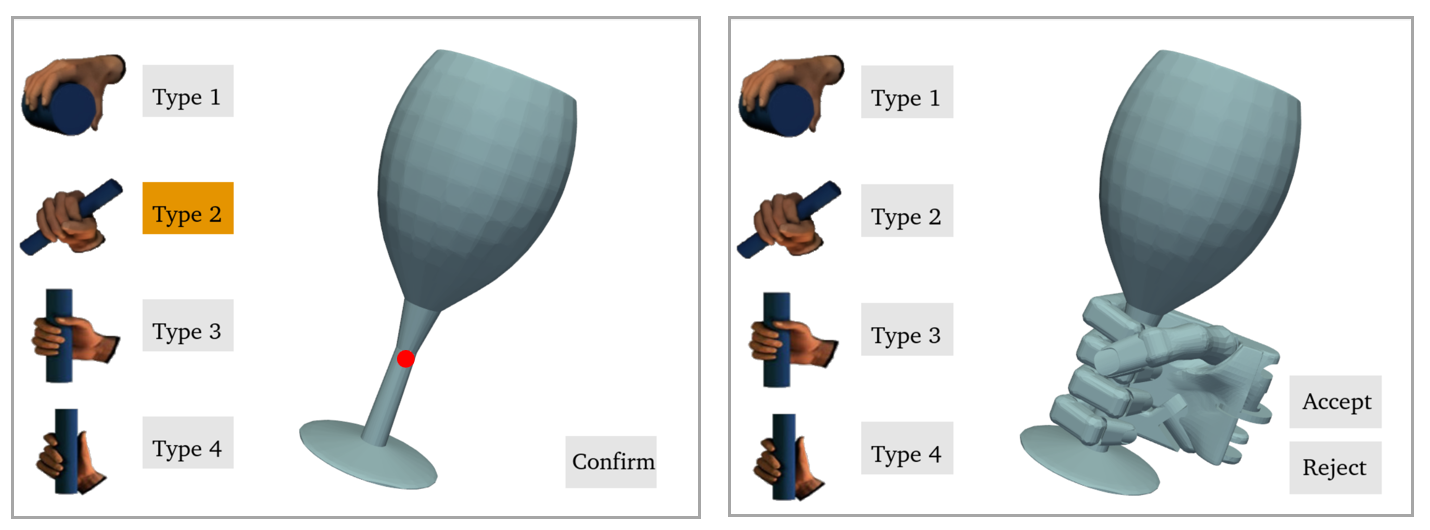}
    \caption{\textbf{An Annotation UI based on Our Algorithm for Collecting Functional Grasp.} (Left) The user \textit{clicks twice} to specify a contact point on the object and a grasp type. (Right) A high-quality grasp is synthesized according to the user's needs within seconds.}
    \label{fig: annotation UI}
\end{figure}

Although our algorithm is semantic-unaware and cannot directly synthesize grasps to touch object regions specified by human language commands, it can be used to develop an efficient annotation system for collecting semantic dexterous grasp data. Unlike widely used teleoperation methods, which often require well-trained annotators and hardware dependencies like data gloves, our annotation system has minimal requirements, relying only on simple mouse clicks.

As shown in Figure~\ref{fig: annotation UI}, the annotator only needs to click twice: once to specify a contact point on the object and once to select a desired grasp type. Our algorithm will automatically sample nearby object points and grasp templates from existing libraries, and synthesize valid grasps, with the best results displayed in the GUI within seconds. For a full demonstration, please refer to our supplementary video. We plan to continue improving this tool and hope it facilitates future research on semantic grasping.

\section{Limitations and Future Work} 
First, our synthesized grasps occasionally fail due to unsuitable or unstable grasp types, as shown in Figure~\ref{fig: real world gallery}. Future work could explore a more suitable taxonomy for robotic grasping, as well as improved strategies for grasp type selection. Second, our method focuses on synthesizing static grasp poses rather than generating sequential trajectories that involve contact changes. A promising direction for future research is to incorporate trajectory generation for dynamic grasping, potentially leveraging reinforcement learning\cite{chen2023visual,yin2025dexteritygen} and differentiable physics simulators~\cite{de2018end, degrave2019differentiable, yang2024jade}. Finally, our work focuses solely on grasping a single object, and the extension to cluttered scenes remains an open challenge for future exploration.

\section{Conclusion} 
\label{sec:conclusion}

In this work, we present a novel pipeline to efficiently synthesize high-quality dexterous grasps for any grasp type, object, and articulated object hand, starting from just one human-annotated template. Our pipeline greatly outperforms previous type-unaware grasp synthesis baselines in the simulation benchmark. Using our proposed method, a large-scale dataset covering 31 grasp types is constructed, enabling the training of a type-conditional grasp generative model. The trained model successfully generates desired grasp types from single-view object point clouds, achieving a real-world success rate of $82.3\%$.

\section*{Acknowledgments}
This work was supported by Beijing Natural Science Foundation (Grant No.QY24042). We also thank Danshi Li for assisting with the integration of the Inspire Hand and the Unitree G1 Hand.

\bibliographystyle{plainnat}
\bibliography{references}

\appendix

\subsection{Post-Filtering for Global Alignment Stage}
\label{app: post filtering}
\begin{figure}[h]
    \centering
    \includegraphics[width=0.9\linewidth]{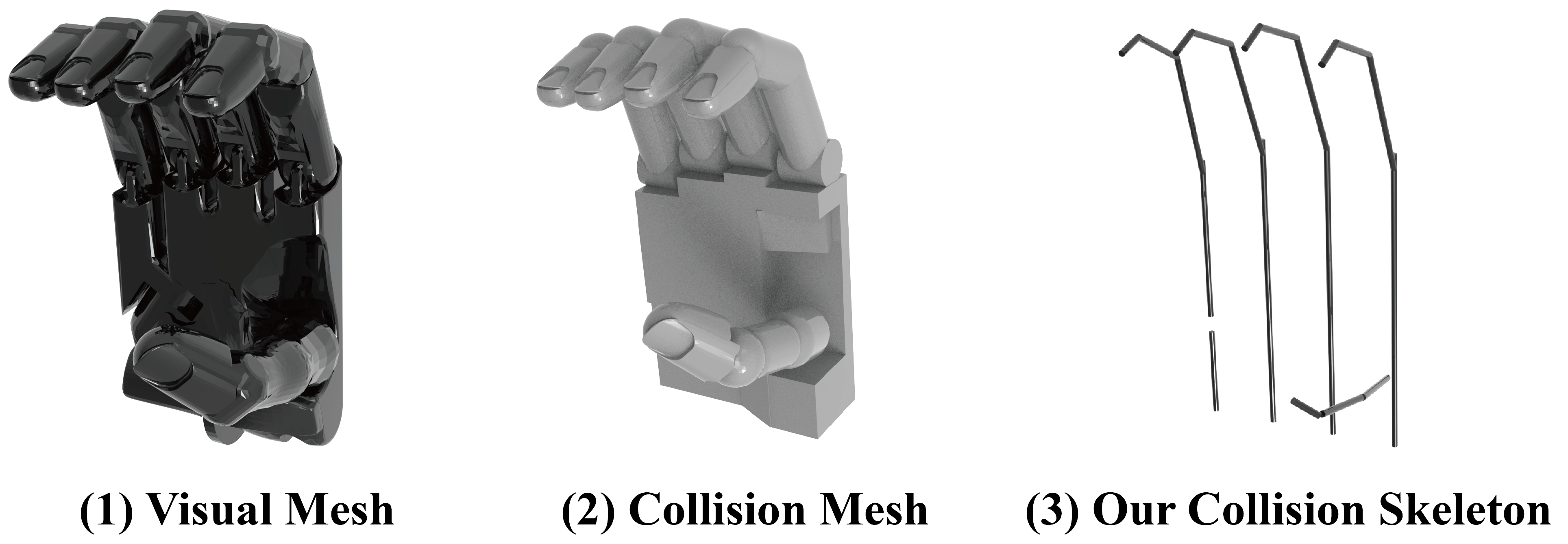}
    \caption{\textbf{Geometric Representations for Shadow Hand.} Our skeleton representation, based on line segments, efficiently detects severe penetrations with the object in the post-filtering of the global alignment stage. The fingertip's skeleton is designed to be shorter, because the fingertip's penetration is often easy to resolve in the local refinement stage.}
    \label{fig: collision skeleton}
\end{figure}

\textbf{Detecting severe penetration}: We propose a skeleton representation for the hand, consisting of several line segments, as shown in Figure~\ref{fig: collision skeleton}. Since the skeleton is parameterized by line segments, collision detection is efficiently performed through an intersection query between the line segment and the object’s mesh. If the posed skeleton collides with the object, the hand is considered to have penetrated too much and is discarded. Note that the skeleton only needs to be defined once for each robotic hand. 

\textbf{Filter out duplicate transformations}: This process iteratively selects the farthest transformations until reaching a predefined number or the farthest distance to other transformations is less than a predefined threshold. The distance metric between two transformations is computed as a weighted sum of: (1) the angular difference between rotations, (2) the Euclidean distance between translations, and (3) optionally, the difference in scales.

\begin{figure*}
    \centering
    \includegraphics[width=0.9\linewidth]{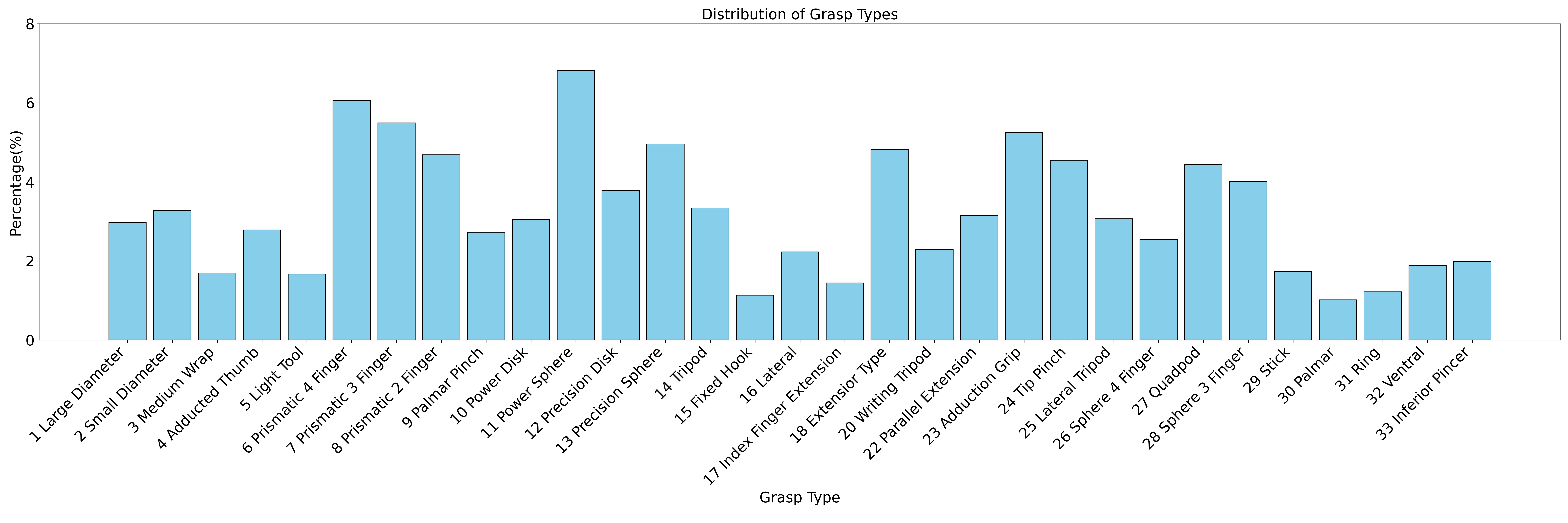}
    \caption{\textbf{Distribution of Grasp Types in the Dexonomy Dataset.} The distribution of grasp types exhibits some bias, which is inevitable due to the varying suitability and stability of different types.}
    \label{fig: grasp type distribution}
\end{figure*}

\subsection{Details of Dexonomy Dataset}
\label{app: dexonomy dataset details}

All objects are pre-processed using CoACD~\cite{wei2022coacd} for convex decomposition, OpenVDB~\cite{museth2013openvdb} for internal face removal, and ACVD~\cite{valette2008generic} for mesh simplification. The OpenVDB and ACVD are essential to support nearest-point queries. Our object pre-processing is very robust, with a success rate over $95\%$ for in-the-wild objects, such as those from Objaverse~\cite{deitke2023objaverse}.

The dataset generation process starts with one manually annotated grasp template per grasp type. Our algorithm runs for 10 epochs, iterating through all objects with 10 attempts per object. Compared to using 1 epoch with 100 attempts, this approach can try more grasp templates for each object, as new templates are actively constructed during the process. 

\begin{figure}
    \centering
    \includegraphics[width=0.9\linewidth]{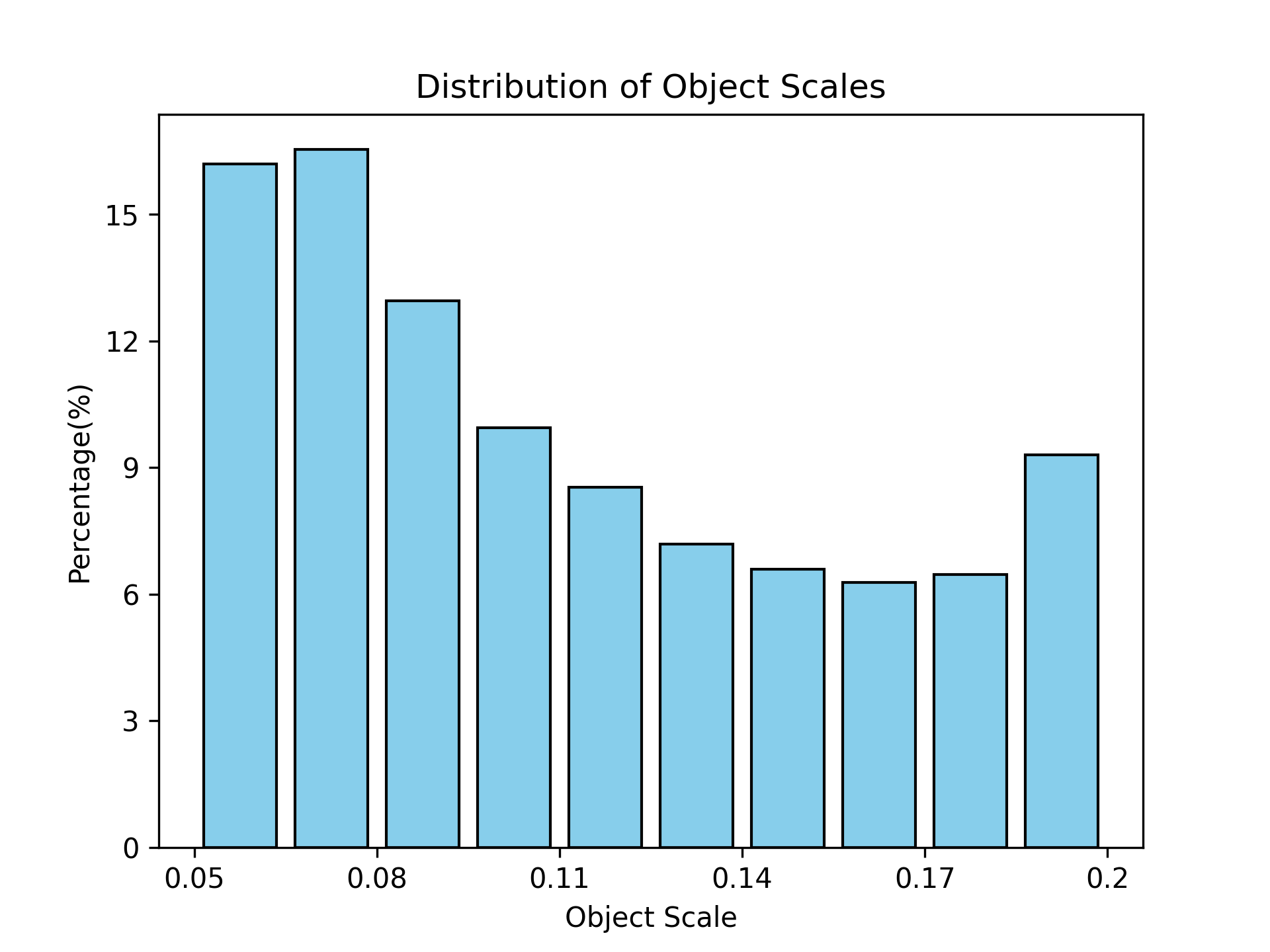}
    \caption{\textbf{Distribution of Object Scales in the Dexonomy Dataset.} The distribution is slightly skewed toward smaller object scales, as smaller objects tend to yield higher grasp success rates—a trend also observed in BODex~\cite{chen2024bodex}.}
    \label{fig: scale distribution}
\end{figure}

The resulting distributions of the object scales and grasp types are shown in Figure~\ref{fig: scale distribution} and Figure~\ref{fig: grasp type distribution}, respectively.

\subsection{Performance Drop of BODex}
\label{app: bodex performance drop}
\begin{figure}
    \centering
    \includegraphics[width=0.9\linewidth]{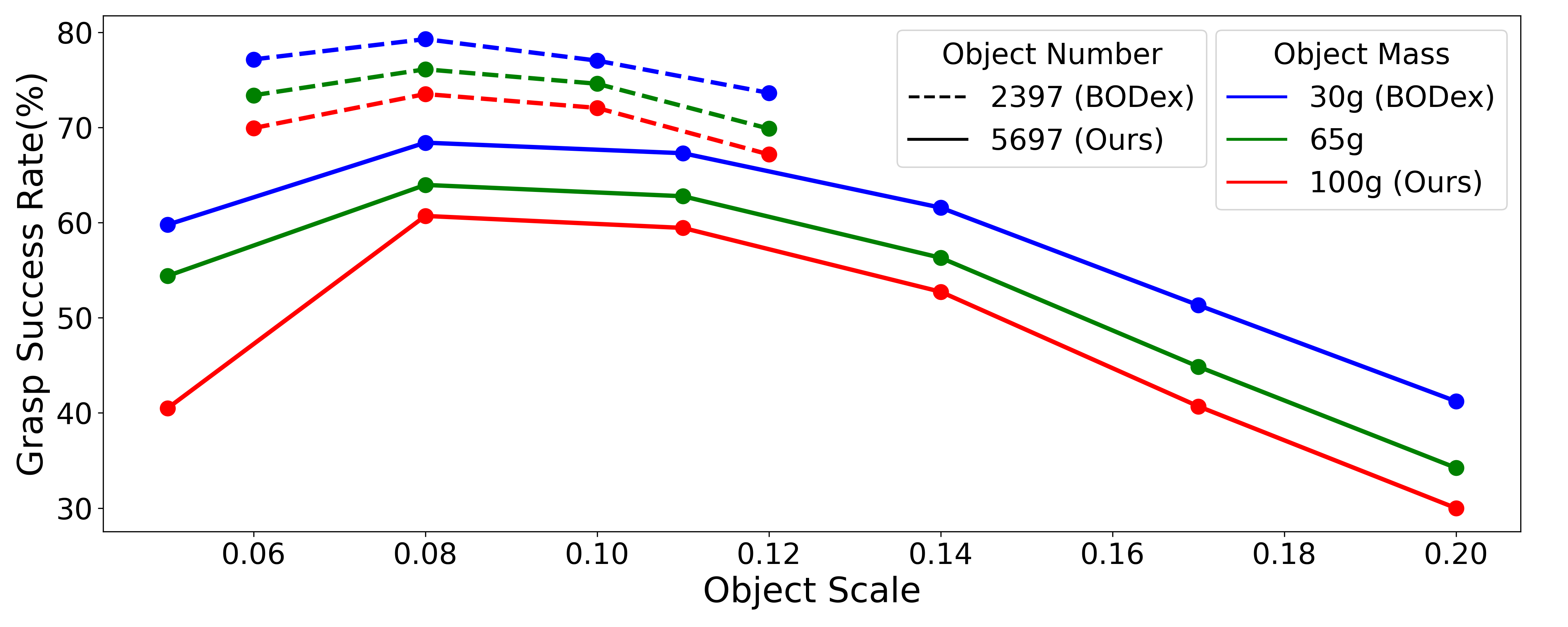}
    \caption{\textbf{Performance Discrepancy of BODex}. The performance of BODex reported in Table~\ref{tab: fingertip baseline} is different from the original paper due to the object set, scale, and mass.}
    \label{fig: bodex performance drop}
\end{figure}

As shown in Figure~\ref{fig: bodex performance drop}, the performance of the type-unaware grasp synthesis baselines in Table~\ref{tab: fingertip baseline} is different from those reported in BODex~\cite{chen2024bodex} because of three reasons:
\begin{itemize}
    \item \textbf{Object set}: BODex selects $2397$ objects from DexGraspNet~\cite{wang2023dexgraspnet} using a heuristic about the minimum AABB length to filter out flat objects, while this work uses all $5697$ objects from DexGraspNet.
    \item \textbf{Object scale}: our paper uses a larger object scale range ($[0.05, 0.2]$) than BODex ($[0.06, 0.12]$).
    \item \textbf{Object mass}: our paper uses a larger mass ($100g$) than BODex ($30g$), which is more challenging.
\end{itemize}

\subsection{More Experimental Results on Learning Method}
\label{app: learning design}

In this section, we present additional preliminary experiments on learning-based grasp synthesis. 

First, we compare the impact of different generative model architectures, as shown in Table~\ref{tab: learning architecture supp}. While diffusion models is more popular~\cite{zhang2024dexgraspnet}, their performance on our dataset is inferior to that of the Mobius Normalizing Flow~\cite{liu2023delving, chen2024bodex}. Additionally, the predicted grasp poses are significantly worse without incorporating an MLP head—a phenomenon also observed in~\cite{zhang2024dexgraspnet}. We hypothesize that this is because the hand joint configurations (qpos) for grasps of the same type are relatively similar and thus easier for an MLP to learn, whereas the 6-DoF root pose of the hand has a more complex distribution that requires the generative model to capture.

Second, we select several grasp types with high success rates from the Dexonomy dataset and train type-unconditional models on them. As shown in Table~\ref{tab: type uncond learning supp}, although the performance varies across grasp types, the overall results remain strong.

Finally, we evaluate different testing strategies for our type-conditional model trained on the full Dexonomy dataset. In the \textit{Average} setting, we generate all 31 grasp types for each testing object point cloud and report the average success rate. In the \textit{Classifier} setting, we train a separate classifier to predict the most suitable grasp type for each object point cloud. \textit{Classifier A} uses the grasp type with the highest success rate per training object in the original Dexonomy dataset as the ground-truth label. In contrast, \textit{Classifier B} synthesizes grasps on training objects using the trained type-conditional model and selects the type with the highest success rate as the ground truth. The classifier employs a 3D sparse convolutional backbone (copied and frozen from the trained type-conditional model) followed by an MLP that outputs grasp type probabilities. The \textit{Top 1} setting directly records the highest success rate across all grasp types for each test object, serving as an oracle upper bound.

As shown in Table~\ref{tab: type cond classifier supp}, the performance of the \textit{Average} strategy is poor, likely because many objects are not compatible with all grasp types. In contrast, the \textit{Top 1} oracle achieves very high performance, indicating the strong potential of our dataset when paired with an effective grasp type selection mechanism. Between the two classifiers, \textit{Classifier B} outperforms \textit{Classifier A}, likely because the grasp type preferences of the trained model differ from those of our pipeline used to synthesize the Dexonomy dataset. Notably, the \textit{Top 1} oracle and using \textit{Classifier B} outperform all type-unconditional models in Table~\ref{tab: type uncond learning supp}, further highlights the advantage of studying different types for grasping.

\begin{table}[]
    \centering
    \begin{tabular}{c|c|c|c|c}
         & Diffusion & Diffusion + MLP & Flow & Flow + MLP \\
         \hline
        GSR ($\%$) $\uparrow$ & 16.0 & 41.8 & NaN & 55.5
    \end{tabular}
    \caption{\textbf{Different Learning Architectures Trained on Grasp Type 1}. \textit{Flow} denotes for mobius normalizing flow. NaN means that the training fails with NaN gradients.}
    \label{tab: learning architecture supp}
\end{table}

\begin{table}[]
    \centering
     \begin{tabular}{c|c|c|c|c|c}
     Dataset & GSR$\uparrow$ & OSR$\uparrow$& CDC$\downarrow$& PD$\downarrow$ & D$\downarrow$ \\
         \hline 
         Ours-type1 &55.5 &85.9 &\textbf{10.8} &8.4 &\textbf{31.5} \\
     Ours-type6 & 48.8
&83.8
&11.4
&7.0
&34.1\\
 Ours-type9 & 48.0
& 79.8
& 15.5
& \textbf{6.6}
& 34.8\\
        Ours-type18 & \textbf{61.2}
&\textbf{87.0}
&13.3
&8.6
&34.0\\
     Ours-type22 & 52.0
&79.1
&12.6
&9.0
&35.4\\
 Ours-type26 & 51.3
& 82.0
&14.0
&10.3
&33.5\\
  Ours-type31 & 47.8
& 81.8
& 14.9
& 6.9
& 32.9\\
Ours-type33 & 46.3
& 77.7
& 15.7
& 6.7
& 33.8\\
    \end{tabular}
    \caption{\textbf{Type-unconditional Model Trained on High-Quality Grasp Types}. The selected grasp types exhibit high success rates during the synthesis of the Dexonomy dataset.}
    \label{tab: type uncond learning supp}
\end{table}

\begin{table}[t]
    \centering
    \begin{tabular}{c|c|c|c|c}
         & Average & Top 1 (oracle) & Classifier A & Classifier B \\
         \hline
        GSR ($\%$) & 28.9 & \textbf{78.2} & 46.1 & 63.9
    \end{tabular}
    \caption{\textbf{Type-Conditional Model with Different Testing Methods.} The performances of \textit{Top 1} oracle and \textit{Classifier B} outperform type-unconditional models in Table~\ref{tab: type uncond learning supp}, highlighting the potential of studying different types for grasping.}
    \label{tab: type cond classifier supp}
\end{table}

\begin{figure}[t]
    \centering
    \includegraphics[width=0.9\linewidth]{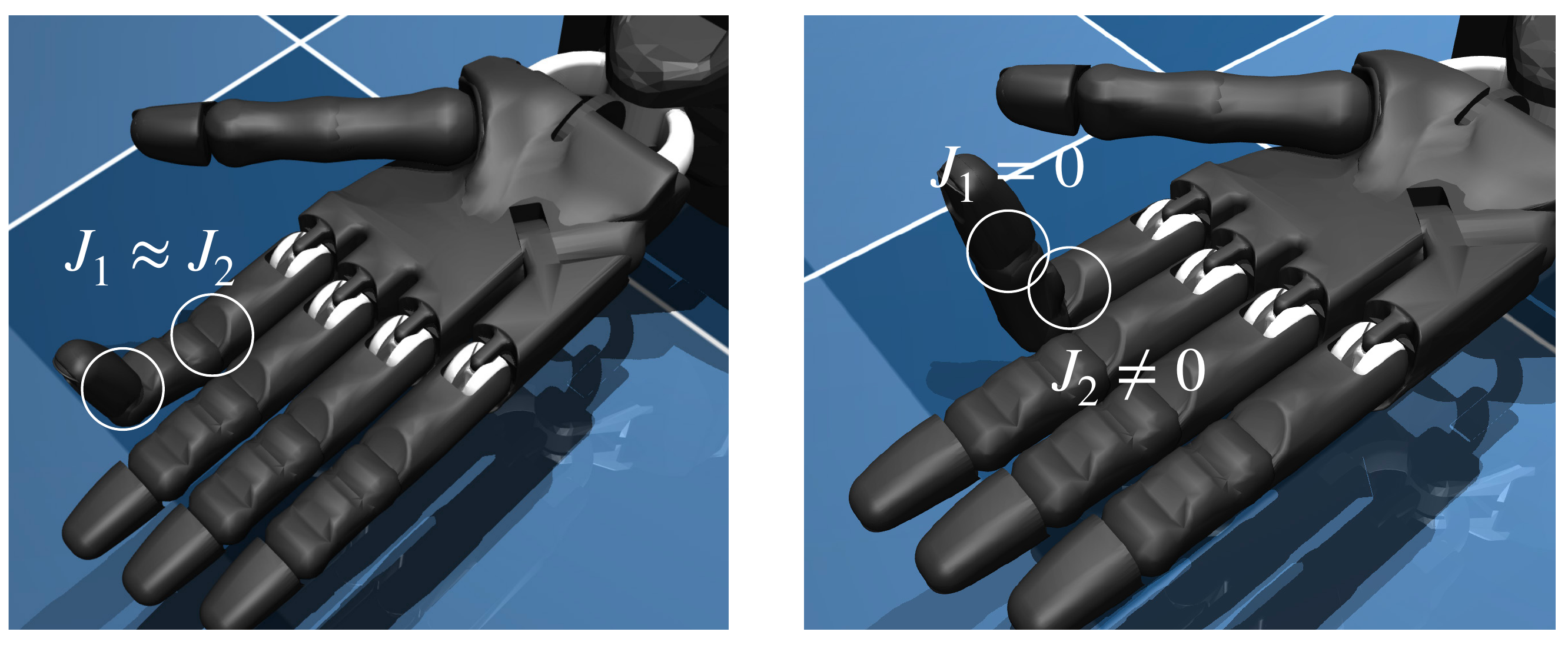}
    \caption{\textbf{Sim-Real Misalignment of Shadow Hand}. (Left) The widely used Shadow hand assets in simulation. (Right) The real-world behavior. }
    \label{fig: real details}
\end{figure}

\subsection{Details of Real-World Experiment}
\label{app: real world details}
One may find that the farthest joint of all fingers, except the thumb finger, never bends in our real-world experiments. This is because the under-actuation mechanism of the real Shadow hand differs from the commonly used one in simulation from MuJoCo Menagerie~\cite{Zakka_MuJoCo_Menagerie_A_2022}, as shown in Figure~\ref{fig: real details}.
In simulation, the two farthest joints of each finger, namely $J_1$ and $J_2$, are always equal to each other. However, in the real world, the farthest joint $J_1$ only bends after the second joint $J_2$ reaches around 90 degrees. At the time of this work, we were unsure how to accurately simulate the phenomenon, so we fixed the farthest joint to zero, annotated new templates, synthesized new data, and trained a new model for real-world deployment. However, we have since found a potential solution and plan to explore it in future work.

Another noticeable issue is related to motion planning. First, the success rate of motion planning is low (less than $30\%$). Second, even when the motion planning succeeds, there are sometimes unexpected collisions while moving to the pre-grasp pose. We suspect that CuRobo may not be well-suited for grasping tasks involving contact, as the planned trajectories are consistently close to the obstacle at all timesteps. Ideally, trajectories should maintain a larger hand-object distance, except in the final few timesteps, which would make them more robust to real-world noise and partial observations. This can be explored in future work. 

\subsection{Detailed Time Analysis}
\label{app: time analysis}

The times reported in Table~\ref{tab: fingertip baseline} represent the maximum speed for synthesis without testing. This section provides a more detailed time breakdown of our proposed grasp synthesis pipeline.

First, the \textit{lightweight global alignment} stage processes over 100,000 initial samples in approximately 3 seconds on a single 3090 GPU. The maximum number of intermediate results generated for the next stage can be controlled via a hyperparameter. We typically process 10 objects in parallel, with 10 results per object. For grasp types that are commonly suitable for many objects, this stage is usually not the bottleneck, as it can synthesize over 200 intermediate results per second using 8 GPUs. However, for more challenging grasp types that are hard to match, this stage can become the bottleneck, because there may be less than 20 results per second.

The optimization step consistently takes around 1.2 seconds, while the time cost of calculating the grasp quality metric for post-filtering varies significantly, ranging from 0.3 to 1.5 seconds. When many samples are filtered out, leaving only around 5,000 for energy calculation, the process takes approximately 0.3 seconds. This speed is achieved by using the batched Relu-QP~\cite{bishop2024relu} algorithm as in BODex~\cite{chen2024bodex}, whereas traditional CPU-based QP solvers are significantly slower. The other operations are very fast.

Next, the \textit{simulation-based local refinement} stage requires 200 simulation steps, which take less than 0.1 seconds. This stage is highly efficient, easily synthesizing more than 200 grasps per second when utilizing 32 threads.

Finally, the \textit{simulation validation} stage often becomes the speed bottleneck, as it involves approximately 3,000 simulation steps (6 external force directions, with 500 steps per direction). Despite employing early-stop strategies to handle failure cases, this stage can only process about 40 grasps per second using 48 threads. The slow speed has nothing to do with our proposed contact-aware control strategy and is consistent for other synthesis baselines if they want to test in MuJoCo. Future work may try to use GPU-based MuJoCo or other faster physics simulators~\cite{li2025taccel} for testing.

\end{document}